\documentclass[]{mc_nankai}

\usepackage[dvipsnames, table, xcdraw]{xcolor}
\usepackage{utfsym}
\usepackage{times}
\usepackage{epsfig}
\usepackage{graphicx}
\usepackage{booktabs}
\usepackage{amsmath, amssymb, mathtools, bm}
\usepackage{multirow, enumitem, subcaption, float, wrapfig}
\usepackage{pifont}
\usepackage{overpic}
\usepackage{tcolorbox}
\usepackage{caption}
\usepackage{tabularx}
\usepackage{algorithm}
\usepackage{algorithmic}
\usepackage{hyperref}
\usepackage{ragged2e}

\def\methodname{PEPO}

\title{Rethinking Token-Level Policy Optimization for Multimodal Chain-of-Thought}

\author[1]{Yunheng Li*}
\author[1]{Hangyi Kuang*}
\author[1]{Hengrui Zhang}

\author[2]{Jiangxia Cao}
\author[2]{Zhaojie Liu}

\author[1]{\\Qibin Hou${\dagger}$}
\author[1]{Ming-Ming Cheng}

\affiliation[1]{VCIP, School of Computer Science, Nankai University}
\affiliation[2]{Kuaishou Technology}

\affiliation[]{$^{*}$Equal Contribution.  $^{\dagger}$Corresponding author.}

\abstract{
Multimodal Chain-of-Thought (CoT) reasoning requires large vision-language models to construct reasoning trajectories that interleave perceptual grounding with multi-step inference.
However, existing Reinforcement Learning with Verifiable Rewards (RLVR) methods typically optimize reasoning at a coarse granularity, treating CoT uniformly without distinguishing their varying degrees of visual grounding.
In this work, we conduct a token-level analysis of multimodal reasoning trajectories and show that successful reasoning is characterized by structured token dynamics reflecting both perceptual grounding and exploratory inference.
Building upon this analysis, we propose \textbf{Perception-Exploration Policy Optimization} (\methodname), which derives a perception prior from hidden state similarity and integrates it with token entropy through a smooth gating mechanism to produce token-level advantages.
\methodname{} integrates seamlessly with existing RLVR frameworks such as GRPO and DAPO, requiring neither additional supervision nor auxiliary branches.
Extensive experiments across diverse multimodal benchmarks demonstrate consistent and robust improvements over strong RL baselines, spanning geometry reasoning, visual grounding, visual puzzle solving, and few-shot classification, while maintaining stable training dynamics.

}

\mclink[Project Page]{\url{https://github.com/xzxxntxdy/PEPO}}

\begin{document}

\maketitle
\justifying

\section{Introduction}
\label{sec:intro}

Large Vision-Language Models (LVLMs)~\cite{bai2025qwen2,zhu2025internvl3,wang2025internvl3.5,hurst2024gpt,team2024gemini,yang2025kwai} have achieved impressive progress across diverse vision-language tasks, such as question answering~\cite{antol2015vqa,goyal2017making}, visual reasoning~\cite{lu2023mathvista,zhang2024mathverse,qiao2024we,qiao2025we2} and reasoning grounding~\cite{kazemzadeh2014referitgame,yu2016modeling,lai2024lisa}.
Recent advances in LVLMs~\cite{shen2025satori,xiao2025advancing,ma2025one,liu2025visionreasoner,liu2025noisyrollout,chen2025vinci,wang2025sota,zhu2025shuffle,yu2025docthinker} have focused on enhancing their reasoning capability, where Reinforcement Learning (RL) serves as an effective way to optimize Chain-of-Thought (CoT)~\cite{wei2022chain} reasoning and improve performance.

\begin{figure*}[tbp]
  \centering
  \includegraphics[width=0.97\textwidth]{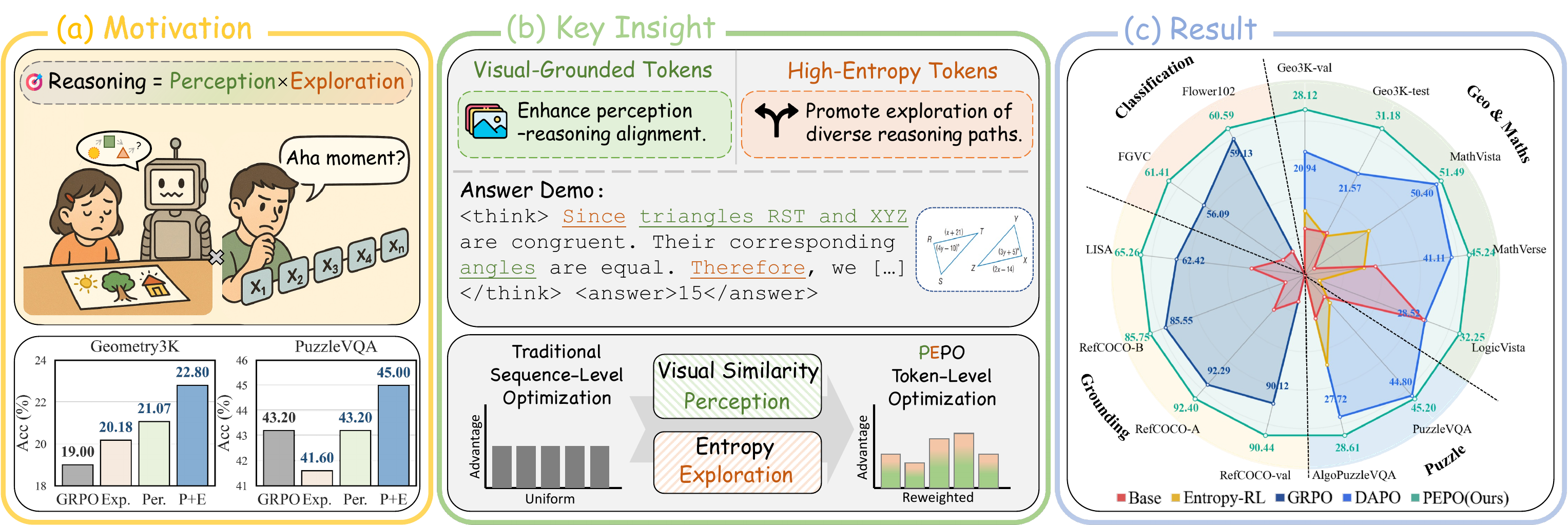}
  \caption{
      Overview of PEPO.
      (a) Effective multimodal reasoning arises from the complementarity between perception and exploration. Abbr. Exp.: Exploration-only, Per.: Perception-only, P+E: Perception + Exploration.
      (b) Unlike traditional sequence-level optimization with uniform advantages, PEPO reweights tokens using a perception prior from visual similarity and entropy via a smooth gate, producing fine-grained token-level advantages.
      (c) When integrated with GRPO or DAPO, PEPO consistently improves performance across diverse benchmarks.
  }
  \label{fig:intro}
\end{figure*}

Typical LVLM training pipelines incorporate RL with verifiable rewards to refine CoT reasoning, commonly under frameworks such as Group Relative Policy Optimization (GRPO)~\cite{guo2025deepseek,shao2024deepseekmath}.
For example, most approaches adopt outcome-based rewards (e.g., format or accuracy) under the assumption that improving answer format or textual correctness naturally leads to coherent reasoning.
However, these methods suffer from sequence-level supervision, which fails to distinguish the contributions of intermediate CoT steps.
To mitigate this, several works in Large Language Models (LLMs) introduce token-level entropy advantages to encourage exploration at uncertain CoT steps~\cite{wang2025harnessing,chen2025seed,wang2025beyond}.
Nevertheless, entropy-based advantages mainly capture textual uncertainty but show weak correspondence to visual semantics and insufficient discrimination of reasoning relevance.
Recent perception-aware RL methods incorporate visual signals, but often introduce additional computational overhead through auxiliary masking branches~\cite{wang2025perception,huang2025spotlight} or attention-based measures~\cite{jian2025look} that are incompatible with efficient acceleration frameworks~\cite{dao2022flashattention}.

Unlike text-only LLMs, LVLMs reason under multimodal constraints, where visual perception and exploratory dynamics play complementary roles in shaping the CoT process, as illustrated in~\figref{fig:intro}(a).
From the perceptual perspective, token-level analysis in~\secref{sec:analysis} reveals that correct reasoning is strongly associated with perceptual grounding: accurate responses consistently depend on a compact subset of visually aligned tokens that anchor the CoT process.
More importantly, a simple hidden-state similarity between response and visual tokens captures this association, providing a modality-specific indicator within the fine-grained reasoning process and reflecting linguistic-perceptual alignment.
Complementing perception, token-level entropy from the logits highlights uncertain steps where alternative reasoning paths should be explored.
However, we find that existing RLVR frameworks overlook the fine-grained coupling between perceptual grounding and reasoning dynamics, relying mainly on outcome- or entropy-based supervision, as well as mask-based perception-aware methods that fail to capture modality-specific interactions.

Motivated by the aforementioned analysis, we introduce Perception-Exploration Policy Optimization (\methodname), a token-level policy optimization framework that couples visual perception and exploration to enhance CoT reasoning in LVLMs.
The core of \methodname~is to convert hidden-state similarity into a calibrated perception prior without auxiliary branches or additional supervision.
Specifically, for each response token, we compute cosine similarity between its hidden state and the set of visual-token states and aggregate it into a per-token visual grounding score.
To incorporate exploration in a unified manner, \methodname~employs a smooth gating mechanism that fuses token-level entropy from the logits with the perception prior to produce a normalized token weight.
These weights refine the sequence-level advantage into token-level advantages, thereby reweighting the policy-gradient updates toward visual-grounded and exploratory reasoning tokens.
Moreover, \methodname~integrates seamlessly with GRPO (\methodname\textsubscript{G}) and DAPO~\cite{yu2025dapo} (\methodname\textsubscript{D}), providing fine-grained optimization signals with only marginal computational overhead.

To validate the effectiveness of \methodname, we evaluate it across multiple multimodal reasoning benchmarks, including geometry and math/logic reasoning, visual puzzles, visual grounding, and few-shot classification.
Across Geometry3K~\cite{lu2021inter}, MathVista-mini~\cite{lu2023mathvista}, MathVerse-mini~\cite{zhang2024mathverse}, and LogicVista~\cite{xiao2024logicvista}, \methodname~improves over GRPO~\cite{guo2025deepseek,shao2024deepseekmath} by +3.67 points on Qwen2.5-VL-3B~\cite{bai2025qwen2} and by +3.51 points on InternVL3-2B~\cite{zhu2025internvl3}, and over DAPO~\cite{yu2025dapo} by +0.45 and +5.15 points, respectively.
On visual puzzles (PuzzleVQA and AlgoPuzzleVQA~\cite{chia2024puzzlevqa}), \methodname~yields gains of +1.65 and +1.52.
For visual grounding (RefCOCO~\cite{yu2016modeling} and LISA-Grounding~\cite{lai2024lisa}), \methodname~achieves +0.86 IoU@50 improvement while avoiding the entropy-only collapse.
In few-shot classification (FGVC Aircraft~\cite{maji2013fgvc} and Flower102~\cite{nilsback2008flower102}), it improves accuracy by +5.32 and +1.46.
Furthermore, scalability analysis on ViRL39k~\cite{wang2025vl} shows that perception-exploration coupling consistent gains with larger data scales, indicating robust generalization and optimization stability across multimodal tasks.
To sum up, our main contributions are threefold:
\begin{itemize}
    \item To our knowledge, this is the first work to explore the complementary roles of visual-grounded and high-entropy tokens in LVLMs, revealing how perception anchors reasoning while entropy drives exploration.
    \item We propose \methodname, a token level policy optimization framework that derives a perception prior from hidden state similarity and incorporates entropy through a smooth gating mechanism to refine advantage estimation.
    \item We instantiate \methodname\textsubscript{G} and \methodname\textsubscript{D} on GRPO and DAPO and obtain consistent gains across geometry, math and logic, visual puzzles, visual grounding, and few-shot classification with marginal overhead.
\end{itemize}

\section{Related Work}

\myPara{RLVR for LVLMs.}
Reinforcement learning with verifiable rewards~\cite{chen2025grpo,shao2024deepseekmath,guo2025deepseek,yu2025dapo} has become an effective approach for training LVLMs.
Among RLVR methods, GRPO is widely used for its stable and critic-free design that directly leverages verifiable rewards for policy optimization.
Recent research has advanced this framework along two main lines of work.
Data-centric studies construct large-scale multimodal datasets and adaptive training schedules to improve generalization~\cite{yang2025r1,liang2025modomodo,meng2025mm,qiao2025we2,wang2025vicrit,ai2025m2,chen2025g1,yuan2025vl,bai2025univg,wang2025internvl3.5,deng2025openvlthinker}.
Meanwhile, reward-centric methods design verifiable rewards for multimodal tasks, such as visual grounding and question answering~\cite{shen2025vlm,liu2025visual,gou2025perceptual,yu2025perception,jiang2025rex,ni2025point,su2025pixel,jiang2025vlm,li2025think,li2025relation,wu2025visualquality}.
Despite these advances, they still rely on sequence-level supervision that overlooks token-level perceptual and reasoning differences.
Recent efforts have explored token-level refinement via entropy-based optimization~\cite{wang2025harnessing,chen2025seed,wang2025beyond,cui2025entropy,vanlioglu2025entropy}, but these methods primarily focus on stabilizing policy updates or enhancing exploration in text-only domains, resulting in limited improvements in LVLM training.

\myPara{Reasoning in LVLMs.}
Reasoning has emerged as a key capability for advancing LVLMs, enabling multi-step inference, numerical computation, and structured visual understanding~\cite{suris2023vipergpt,chen2023shikra,peng2023kosmos}.
Existing approaches enhance reasoning in LVLMs through chain-of-thought supervision and step-wise instruction tuning~\cite{xu2024llavacot,zhang2025improve,mitra2024compositional}, which encourage structured inference but remain limited by static supervision and lack adaptive feedback.
To address this, reinforcement learning has been employed to refine reasoning consistency and correctness~\cite{wang2025vl,wan2025srpo,fan2025sophiavl,chen2025grpo}, introducing dynamic optimization signals for reasoning refinement.
Recent RL-based studies further incorporate verifiable or task-specific rewards for logical reasoning, mathematical derivation, and spatial problem solving~\cite{tan2025reason,li2025think,shen2025satori,xiao2025advancing,ma2025one,liu2025visionreasoner}.
Meanwhile, an emerging line of work operationalizes perception as tool use, employing visual operations such as cropping and zooming~\cite{wu2025reinforcing,zhang2025chain,sarch2025grounded,xu2025visual,su2025openthinkimg,zheng2025deepeyes,fan2025grit,zhu2025active}.
Nevertheless, existing RL-based reasoning frameworks mainly optimize textual consistency while insufficiently leveraging visual perception and exploratory dynamics that are essential for multimodal reasoning.

\section{Methodology}

\subsection{Background and Motivation}
\label{sec:background}

GRPO~\cite{shao2024deepseekmath} has become a widely used RL algorithm to enhance the reasoning capabilities of large language and vision-language models.
It performs policy optimization through group-wise relative evaluation, where multiple responses are sampled for each query and their verifiable rewards are compared to obtain advantages for policy updates.
Specifically, for each input query, GRPO generates a group of $G$ candidate responses and evaluates them using verifiable rewards $\{R^{(i)}\}_{i=1}^{G}$ to enable reward-based comparison within the group.
From these evaluations, the advantage of the $i$-th response is defined as:
\begin{equation}
A^{(i)} = \frac{R^{(i)} - \text{mean}({R^{(j)}})}{\text{std}({R^{(j)}})},
\end{equation}
which represents its relative reward among the sampled responses.
During training, this advantage $A^{(i)}$ is uniformly applied to all tokens of the $i$-th response, and the policy parameters are updated using a PPO-style~\cite{schulman2017proximal} objective:
\begin{equation}
J_{\text{G}}(\theta) =
\mathbb{E}\!\left[\min\!\big(r_{t}^{(i)}A^{(i)},\,
\mathrm{clip}(r_{t}^{(i)},1-\varepsilon,1+\varepsilon)A^{(i)}\big)\right],
\end{equation}
where $r_{t}^{(i)}$ denotes the importance ratio computed from the new and old policies, and $\varepsilon$ is the clipping threshold that controls the update magnitude for stable policy optimization.
From this objective, the policy gradient in GRPO is driven by the sequence-level advantage $A^{(i)}$, which is uniformly applied across all tokens within the response.

However, such sequence-level supervision limits optimization granularity, as it ignores the varying semantic and perceptual relevance of individual tokens.
This limitation is particularly pronounced in large vision-language models, where visual grounding primarily determines response correctness, whereas textual reasoning contributes more extensively to gradient updates, leading to optimization imbalance and weakened perception-reasoning alignment.

\begin{figure*}[tbp]
  \centering
  \begin{subfigure}{0.32\linewidth}
    \centering
    \includegraphics[width=\linewidth]{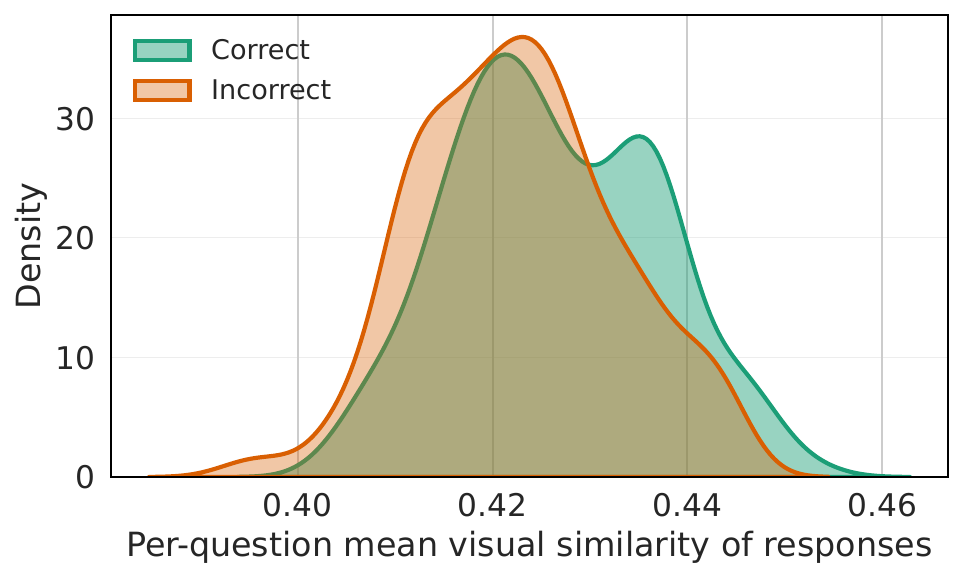}
    \caption{Global similarity ($M_{\text{glob}}$)}
    \label{fig:overall_mean}
  \end{subfigure}
  \begin{subfigure}{0.32\linewidth}
    \centering
    \includegraphics[width=\linewidth]{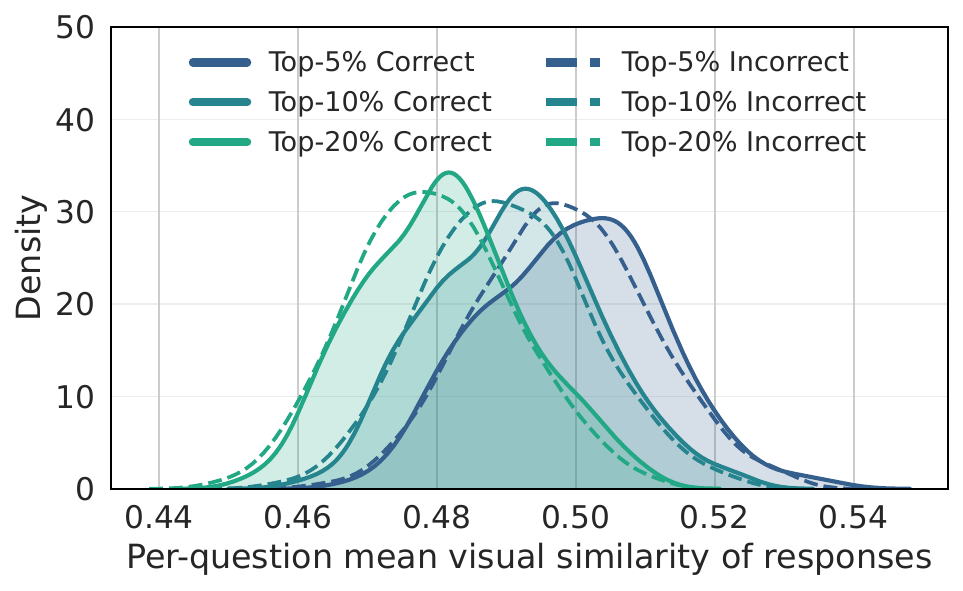}
    \caption{Top-$K$ similarity ($M_{\text{high}}$)}
    \label{fig:topk_mean}
  \end{subfigure}
  \begin{subfigure}{0.32\linewidth}
    \centering
    \includegraphics[width=\linewidth]{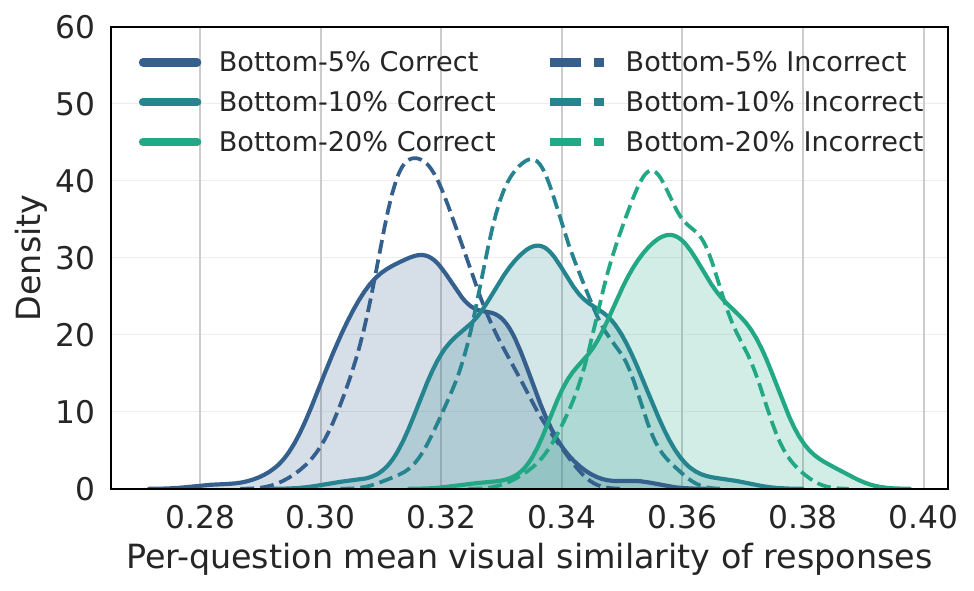}
    \caption{Bottom-$K$ similarity ($M_{\text{low}}$)}
    \label{fig:bottomk_mean}
  \end{subfigure}
  \caption{
  Distributions of different visual similarity metrics comparing correct and incorrect responses.
  (a) The global similarity ($M_{\text{glob}}$) across all tokens, where correct responses exhibit a clear rightward shift.
  (b) The top-$K$ similarity ($M_{\text{high}}$), where the correct-response peak also moves right.
  (c) The bottom-$K$ similarity ($M_{\text{low}}$), where the shift is negligible.
  Together, these results show that reasoning correctness is characterized by a subset of visual-grounded tokens.}
  \label{fig:three_horizontal}
\end{figure*}

\subsection{Token-Level Analysis of Multimodal Reasoning}
\label{sec:analysis}

To investigate how token-level signals relate to multimodal reasoning behavior, we analyze visual similarity and entropy as complementary indicators of perceptual grounding and reasoning uncertainty.
Our analysis is conducted on the Geometry3K dataset~\cite{lu2021inter} using the Qwen2.5-VL-3B-Instruct model~\cite{bai2025qwen2}, sampling 8 responses per question with a decoding temperature of 1.

\myPara{Visual similarity analysis.}
To quantify the visual dependency of each response token, we define its visual similarity (VS) as the mean cosine similarity between the hidden states of the response token and those of all vision tokens across all model layers:
\begin{equation}
\mathrm{VS}_t = \frac{1}{L}\!\sum_{l=1}^{L}\frac{1}{N}\!\sum_{n=1}^{N}
\frac{\langle h_{l,t}, v_{l,n}\rangle}{\|h_{l,t}\|\|v_{l,n}\|},
\end{equation}
where $L$ denotes the total number of layers, $N$ the number of vision tokens, and $h_{l,t}$ and $v_{l,n}$ represent the hidden states of the $t$-th response token and the $n$-th vision token at layer $l$, respectively.

To assess how visual dependency correlates with response correctness, we aggregate the $\mathrm{VS}$ scores within each response across all tokens ($M_{\text{glob}}$), the top-$K$ subset ($M_{\text{high}}$), and the bottom-$K$ subset ($M_{\text{low}}$). For each question, these metrics are computed separately for correct and incorrect responses.
As shown in \figref{fig:three_horizontal}, the distributions of $M_{\text{glob}}$ and $M_{\text{high}}$ for correct responses exhibit a clear rightward shift relative to incorrect ones, indicating that successful reasoning places greater weight on a compact subset of visually aligned tokens.
In contrast, the $M_{\text{low}}$ distributions show minimal separation, suggesting that tokens with low visual relevance contribute little to distinguishing response quality.
These observations suggest that correctness is associated with increased reliance on visual-grounded tokens.

\begin{figure*}[t]
  \centering
  \includegraphics[width=0.98\textwidth]{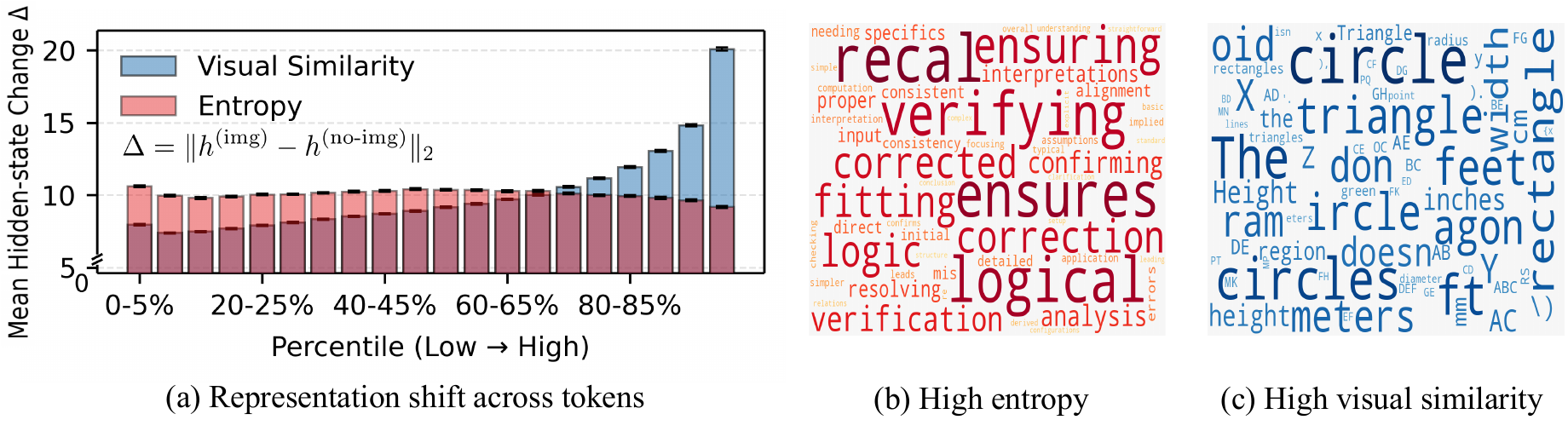}
  \caption{
    Token-level analysis of visual similarity and entropy.
    (a) High visual similarity tokens exhibit larger hidden-state shifts under image removal than high entropy tokens.
    (b) Word cloud of high entropy tokens and (c) word cloud of high visual similarity tokens, illustrating reasoning-related and perceptual terms.
  }
  \label{fig:representation_shift}
\end{figure*}

\myPara{Visual-entropy complementarity.}
To assess the complementary contributions of visual similarity and entropy, we analyze token partitions defined by these two indicators under controlled perturbations and through their associated semantic patterns.
As shown in Fig.~\ref{fig:representation_shift}(a), we conduct a controlled forward pass using identical question-response pairs while removing the image input.
Tokens ranked by visual similarity exhibit substantially larger hidden-state shifts under image removal, indicating stronger dependence on visual evidence.
In contrast, entropy-ranked tokens show relatively stable representations, suggesting that entropy primarily reflects language uncertainty rather than visual sensitivity.
We further analyze the token distributions associated with each partition.
Fig.~\ref{fig:representation_shift}(b) shows that high-entropy tokens are enriched with reasoning-transition expressions such as verification, correction, and analysis, which typically mark decision points within the reasoning trajectory.
In comparison, Fig.~\ref{fig:representation_shift}(c) illustrates that high visual similarity tokens concentrate on perceptually grounded concepts, including geometric entities and spatial attributes.
These observations indicate that visual similarity and entropy encode complementary aspects of multimodal reasoning: the former reflects perceptual grounding, whereas the latter quantifies uncertainty throughout the reasoning process.

\subsection{Perception-Exploration Policy Optimization}
\label{sec:method}

\begin{figure*}[t]
  \centering
  \includegraphics[width=0.95\textwidth]{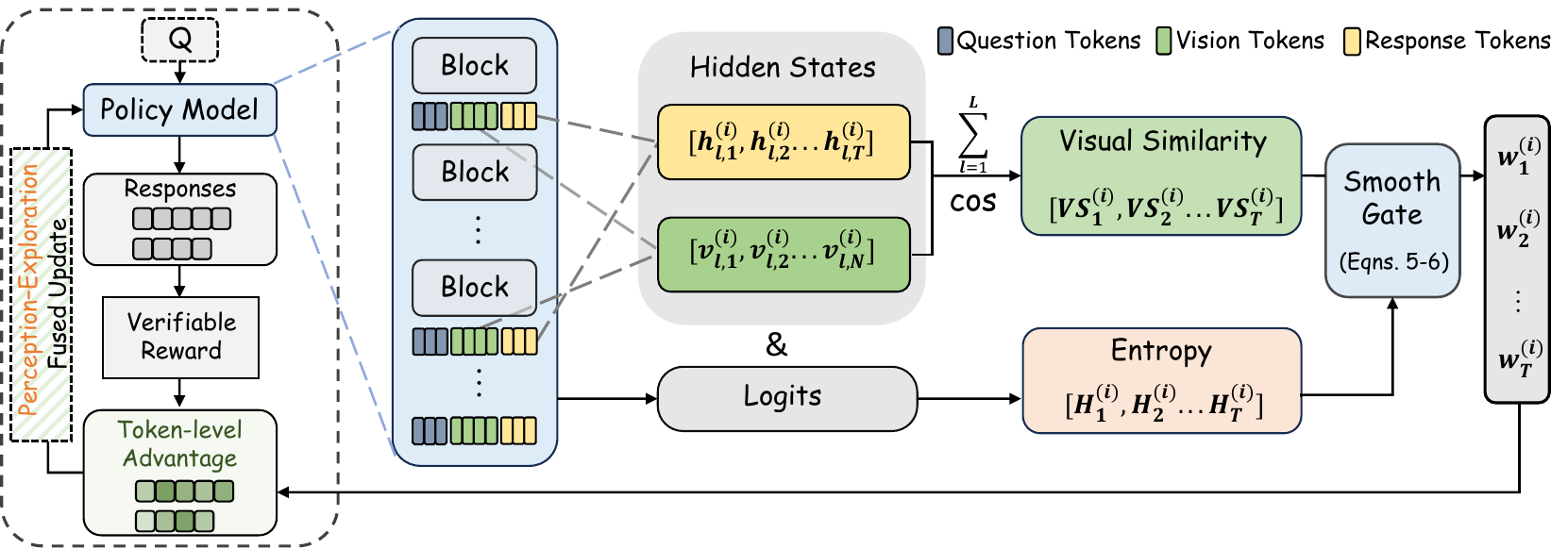}
  \caption{
    Framework of PEPO.
    During response generation, the layer-wise hidden states of response tokens and vision tokens are extracted, along with the output logits.
    For each response token, visual similarity and entropy are computed, and the centered sum of their normalized values is passed through a smooth gating function to produce token-wise weights that modulate the advantages for PEPO updates.
  }
  \label{fig:framework}
\end{figure*}

Building upon the above analysis, we introduce Perception-Exploration Policy Optimization (\methodname), a token-level reinforcement learning framework that integrates visual perception and exploration to refine reasoning in LVLMs.
As illustrated in~\figref{fig:framework}, \methodname~extracts layer-wise hidden states of response and vision tokens from the policy model and computes visual similarity and entropy for each token to capture perceptual grounding and reasoning uncertainty.
Our key insight is that visual-grounded tokens anchor perception and high-entropy tokens capture exploratory transitions, which together exhibit complementary roles during multimodal reasoning.
To exploit this complementarity, \methodname\ employs a smooth gating mechanism that fuses visual similarity and entropy to generate adaptive token-level weights.
These weights induce token-level advantages that reweight the policy-gradient updates toward visual-grounded and exploratory reasoning tokens, enabling fine-grained optimization that distinguishes the contributions of intermediate reasoning steps.

\myPara{Perception modeling.}
Based on the analysis in~\secref{sec:analysis} that visual-grounded tokens play a key role in reasoning accuracy, we incorporate a perception prior $\{\mathrm{VS}_t^{(i)}\}_{t=1}^{T}$ to capture the degree of visual grounding for each token in the $i$-th response.
To this end, each $\mathrm{VS}_t^{(i)}$ is computed from the hidden-state correlations between response and vision tokens across all transformer layers, serving as a lightweight and supervision-free estimate of perceptual alignment.
This design allows tokens with higher $\mathrm{VS}_t^{(i)}$ to receive greater importance during optimization, thereby guiding the model to focus on visual-grounded reasoning steps.

\myPara{Exploration modeling.}
While the perception modeling captures how strongly each token is grounded in visual content, reasoning dynamics in LVLMs also involve uncertainty and exploratory transitions that perception alone cannot represent.
To model this aspect, we introduce an exploration score computed from the token-level entropy sequence $\{H_t^{(i)}\}_{t=1}^{T}$ derived from the output logits of the policy model:
\begin{equation}
H_t^{(i)} = -\sum_{x \in \mathcal{V}} p_\theta(x|s_t^{(i)})\log p_\theta(x|s_t^{(i)}),
\end{equation}
where $\mathcal{V}$ denotes the vocabulary and $p_\theta(x|s_t^{(i)})$ is the model's predicted probability of token $x$ at decoding state $s_t^{(i)}$.
Tokens with higher $H_t^{(i)}$ correspond to uncertain reasoning steps or transition points, reflecting regions where the model explores multiple reasoning responses.
By integrating this exploration signal with perception modeling, \methodname~achieves a more fine-grained representation of multimodal reasoning processes.

\myPara{Perception-exploration fusion.}
To construct a unified optimization framework, we integrate the perception score $\mathrm{VS}_t^{(i)}$ and exploration score $H_t^{(i)}$ to jointly model perceptual grounding and exploratory uncertainty at the token level.
Both scores are min-max normalized to [0,1] within each response to obtain $\hat{\mathrm{VS}}_t^{(i)}$ and $\hat{H}_t^{(i)}$, ensuring comparability and preventing scale bias, and their joint dependency is parameterized through a smooth gating operator:
\begin{align}
\hat{g}_t^{(i)} &= \hat{\mathrm{VS}}_t^{(i)} + \hat{H}_t^{(i)} - \mathrm{mean}_t(\hat{\mathrm{VS}}^{(i)} + \hat{H}^{(i)}), \label{eq:centered-sum}\\
w_t^{(i)} &= T \cdot \mathrm{Softmax}\!\big((1 + \alpha \tanh(\hat{g}_t^{(i)})) \cdot \mathrm{VS}_t^{(i)}\big),
\label{eq:joint-gate}
\end{align}
where $\hat{g}_t^{(i)}$ denotes the mean-centered joint score derived from the normalized visual similarity and entropy, which is subsequently processed by a $\tanh(\cdot)$ activation to obtain a smooth gating function.
Crucially, the gate is multiplied by $\mathrm{VS}_t^{(i)}$, which keeps perception dominant and conditions entropy-driven modulation on visually grounded tokens, avoiding indiscriminate amplification of high-entropy but visually irrelevant tokens.
Finally, $T$ rescales the softmax output so that $\mathbb{E}[w_t^{(i)}]=1$, preserving the overall advantage scale while redistributing token-level credit.
This operator adaptively assigns token-level weights $w_t$ by integrating perceptual relevance and reasoning uncertainty, which subsequently guide token-level policy optimization in a fine-grained manner.

\myPara{Token-level advantage.}
The fused weight $w_t^{(i)}$ is used to refine the sequence-level advantage computed in GRPO variants.
Let $A^{(i)}$ denote the GRPO advantage for the $i$-th response.
We define a token-level advantage as:
\begin{equation}
A_t^{(i)} = \big[(1-\lambda) + \lambda w_t^{(i)}\big]A^{(i)},
\end{equation}
where $\lambda$ controls the strength of the token-level modulation, linearly increasing from 0 to 1 over training steps.
This formulation yields fine-grained optimization signals that capture the heterogeneous contributions of individual tokens, allowing \methodname~to be seamlessly incorporated into policy optimization frameworks while preserving computational efficiency and enhancing perception-reasoning alignment.

\section{Experiments}
\label{sec: setting}

\subsection{Experiment Setup}

\myPara{Models and baselines.}
We conduct experiments using two recent open-source vision-language models, Qwen2.5-VL-3B-Instruct~\cite{bai2025qwen2} and InternVL3-2B-Instruct~\cite{zhu2025internvl3}, both of which exhibit strong multimodal representation and reasoning capabilities.
To evaluate effectiveness, \methodname~is compared against three representative RL methods: GRPO~\cite{shao2024deepseekmath}, DAPO~\cite{yu2025dapo}, and High-Entropy RL~\cite{wang2025beyond}.
Since \methodname~introduces token-level advantage estimation, it remains fully compatible with existing policy optimization frameworks.
Two variants are accordingly implemented: \methodname\textsubscript{G}, built upon GRPO, and \methodname\textsubscript{D}, built upon DAPO.

\myPara{Datasets.}
We evaluate \methodname~ across five categories of multimodal reasoning tasks.
For \textbf{geometry reasoning}, training is conducted on Geometry3K~\cite{lu2021inter}, and generalization is assessed on MathVista~\cite{lu2023mathvista}, MathVerse~\cite{zhang2024mathverse}, and LogicVista~\cite{xiao2024logicvista}, reporting average accuracy over 8 responses (avg@8).
For \textbf{visual grounding}, we use 2K samples from RefCOCO for RL training and evaluate on the validation, testA, and testB splits, with cross-domain evaluation on LISA-Grounding~\cite{lai2024lisa} using IoU@50.
For \textbf{few-shot classification}, we adopt FGVC Aircraft~\cite{maji2013fgvc} and Flower102~\cite{nilsback2008flower102} under 1-, 2-, and 4-shot settings to examine data efficiency.
For \textbf{visual puzzle reasoning}, we randomly sample 1.5K examples from the PuzzleVQA dataset~\cite{chia2024puzzlevqa} for training and reserve 0.5K for testing, with AlgoPuzzleVQA used for out-of-domain evaluation.
Finally, for \textbf{scalability analysis}, we train on the large-scale ViRL39K dataset~\cite{wang2025vl} and evaluate on diverse reasoning benchmarks including Geometry3K\textsubscript{test}, MathVista, We-Math~\cite{qiao2024we}, MathVerse, LogicVista, SuperClevr Counting~\cite{li2023superclevr}, and MMMU-Pro~\cite{yue2025mmmu}.

\myPara{Implementation details.}
All models are trained under a unified RLVR framework using the default verifiable rewards defined by each dataset.
The hyperparameter $\alpha$ is tuned separately for each dataset due to task differences.
We use AdamW~\cite{loshchilov2017decoupled} as the optimizer with full-parameter fine-tuning, bfloat16 precision, and gradient checkpointing enabled to reduce memory consumption.
For each question, we sample 8 responses using a temperature of 1.0 and top-$p = 1.0$, and train with DeepSpeed ZeRO-2~\cite{rasley2020deepspeed} for efficient distributed optimization.
All RLVR methods are implemented within the Swift framework~\cite{zhao2025swift} and all experiments are conducted on 8 NVIDIA A40 GPUs.
Additional implementation details are provided in the supplementary material.

\begin{table*}[t]
  \centering
  \setlength{\tabcolsep}{4.5pt}
  \setlength{\abovecaptionskip}{0pt}
  \caption{
  Results on Geometry3K validation/test splits and out-of-domain benchmarks, including MathVista, MathVerse, and LogicVista.
  }
  \label{tab:geom3k_acc}
  \resizebox{\linewidth}{!}{
  \begin{tabular}{lcccccc}
  \toprule
  \textbf{Method} & \textbf{Geometry3K}\textsubscript{val} & \textbf{Geometry3K}\textsubscript{test} & \textbf{MathVista\textsubscript{mini}} & \textbf{MathVerse\textsubscript{mini}} & \textbf{LogicVista} & \textbf{Avg} \\
  \midrule
  \multicolumn{7}{l}{\textit{Qwen2.5-VL-3B-Instruct}} \\
  \addlinespace[2pt]
  Base (zero-shot)      & 11.58 & 14.25 & 50.19 & 37.66 & 32.22 & 29.18 \\
  High-Entropy RL       & 15.58 & 18.26 & 50.69 & 39.84 & 32.59 & 31.39 \\
  GRPO                  & 19.00 & 23.79 & 51.56 & 40.54 & 28.30 & 32.64 \\
  \rowcolor[HTML]{EFEFEF} \methodname\textsubscript{G} (Ours) & 21.91 & 27.27 & 54.45 & 45.42 & 34.45 & 36.70~\textcolor{blue}{(+4.06)}\\
  DAPO                  & 22.63 & 27.00 & 53.68 & 44.44 & 33.13 & 36.18 \\
  \rowcolor[HTML]{EFEFEF} \methodname\textsubscript{D} (Ours) & 22.38 & 27.89 & 54.75 & 44.23 & 33.89 & 36.63~\textcolor{blue}{(+0.45)} \\
  
  \midrule
  \multicolumn{7}{l}{\textit{InternVL3-2B-Instruct}} \\
  \addlinespace[2pt]
  Base (zero-shot)      & 7.88  & 9.03  & 23.00 & 23.36 & 28.36 & 18.33 \\
  High-Entropy RL       & 10.89 & 8.30  & 35.24 & 20.70 & 17.18 & 18.46 \\
  GRPO                  & 22.08 & 26.33 & 49.36 & 42.09 & 28.84 & 33.74 \\
  \rowcolor[HTML]{EFEFEF} \methodname\textsubscript{G} (Ours) & 25.84 & 31.28 & 52.36 & 44.89 & 31.88 & 37.25~\textcolor{blue}{(+3.51)} \\
  DAPO                  & 20.94 & 21.57 & 50.40 & 41.11 & 28.52 & 32.51 \\
  \rowcolor[HTML]{EFEFEF} \methodname\textsubscript{D} (Ours) & 28.12 & 31.18 & 51.49 & 45.24 & 32.25 & 37.66~\textcolor{blue}{(+5.15)} \\
  \bottomrule
  \end{tabular}
  }
\end{table*}

\begin{table}[t]
\centering
\small

\begin{minipage}[t]{0.5\linewidth}
  \captionsetup{type=table}
  \setlength{\tabcolsep}{2pt}
  \setlength{\abovecaptionskip}{2pt}
  \caption{IoU@50 on RefCOCO and out-of-domain LISA with Qwen2.5-VL-3B-Instruct. The High-Entropy RL collapses in 3 runs, so its results are omitted.}
  \label{tab:RefCOCO_LISA_iou}
  \centering
  \resizebox{\linewidth}{!}{%
    \begin{tabular}{lcccccc}
    \toprule
    \multirow{2}{*}{\textbf{Method}}
    & \multicolumn{3}{c}{\textbf{RefCOCO}}
    & \multirow{2}{*}{\textbf{LISA}}
    & \multirow{2}{*}{\textbf{Avg}} \\
    \cmidrule(lr){2-4}
    & val & testA & testB &  &  \\
    \midrule
    Base (zero-shot)      & 89.10 & 91.70 & 84.00 & 56.51 & 80.33 \\
    High-Entropy RL       & \multicolumn{5}{c}{model collapsed (no valid results)} \\
    GRPO                  & 90.12 & 92.29 & 85.55 & 62.42 & 82.60 \\
    \rowcolor[HTML]{EFEFEF} \methodname\textsubscript{G} (Ours) & 90.44 & 92.40 & 85.75 & 65.26 & 83.46~\textcolor{blue}{(+0.86)} \\
    DAPO                  & 90.10 & 92.40 & 85.08 & 62.48 & 82.52 \\
    \rowcolor[HTML]{EFEFEF} \methodname\textsubscript{D} (Ours) & 90.03 & 92.72 & 85.42 & 64.23 & 83.10~\textcolor{blue}{(+0.58)} \\
    \bottomrule
    \end{tabular}%
  }
\end{minipage}\hfill
\begin{minipage}[t]{0.48\linewidth}
  \captionsetup{type=table}
  \setlength{\tabcolsep}{2pt}
  \setlength{\abovecaptionskip}{2pt}
  \renewcommand{\arraystretch}{1.15}
  \caption{Few-shot (1/2/4-shot) results on the FGVC Aircraft and Flower102 datasets. All methods are trained and evaluated using the Qwen2.5-VL-3B-Instruct backbone.}
  \label{tab:fewshot_combined}
  \centering
  \resizebox{\linewidth}{!}{%
    \begin{tabular}{lccccc}
    \toprule
    \textbf{Dataset} & \textbf{Method} & \textbf{1-shot} & \textbf{2-shot} & \textbf{4-shot} & \textbf{Avg} \\
    \midrule
    \multirow{3}{*}{FGVC}
    & SFT & 36.24 & 40.86 & 55.48 & 44.19 \\
    & GRPO & 48.72 & 55.60 & 63.94 & 56.09 \\
    \rowcolor[HTML]{EFEFEF} & \methodname\textsubscript{G} & 51.13 & 57.31 & 75.79 & 61.41~\textcolor{blue}{(+5.32)} \\
    \midrule
    \multirow{3}{*}{Flower102}
    & SFT & 33.98 & 44.50 & 52.33 & 43.60 \\
    & GRPO & 53.23 & 57.94 & 66.22 & 59.13 \\
    \rowcolor[HTML]{EFEFEF} & \methodname\textsubscript{G} & 54.81 & 59.44 & 67.52 & 60.59~\textcolor{blue}{(+1.46)} \\
    \bottomrule
    \end{tabular}%
  }
\end{minipage}

\end{table}

\subsection{Main Results}
\label{sec:main_results}

\myPara{Overall performance.}
Across all benchmarks and model architectures, \methodname~consistently outperforms GRPO, DAPO, and High-Entropy RL in both reasoning accuracy and training stability.
Compared with GRPO, High-Entropy RL shows unstable optimization and weaker multimodal reasoning performance, as it relies solely on entropy-driven exploration without perceptual grounding.
These results show that visual grounding complements entropy-based exploration in achieving robust multimodal reasoning.
By integrating visual similarity to strengthen perception and token entropy to guide exploration, \methodname~achieves stable optimization and notable performance gains across tasks.
We next present detailed evaluations on geometry reasoning, visual grounding, and few-shot classification, followed by results on visual puzzle reasoning.

\myPara{Geometry reasoning.}
\tabref{tab:geom3k_acc} presents results on Geometry3K and three out-of-domain benchmarks, MathVista, MathVerse, and LogicVista.
On Qwen2.5-VL-3B, \methodname\ improves the average score over GRPO by +3.67 points and over DAPO by +0.45 points.
On InternVL3-2B, the improvements over GRPO and DAPO are +3.51 and +5.15 points.
Larger performance gains on MathVerse and LogicVista indicate that \methodname\ is effective on benchmarks requiring integrated visual and symbolic reasoning.

\myPara{Visual grounding.}
\tabref{tab:RefCOCO_LISA_iou} presents results on RefCOCO and the out-of-domain LISA-Grounding dataset evaluated by IoU@50.
On RefCOCO, \methodname~achieves comparable or slightly higher accuracy than GRPO and DAPO across all validation and test splits, indicating stable in-domain performance.
More evident improvements are observed on LISA-Grounding, where \methodname~shows clear gains in localization accuracy under domain shift.
These results suggest that weighting tokens by visual similarity improves the alignment between textual and visual representations.

\begin{table*}[t]
  \small
  \setlength{\tabcolsep}{8.2pt}
  \setlength{\abovecaptionskip}{2pt}
  \centering
  \caption{
  Accuracy (\%) on PuzzleVQA and out-of-domain AlgoPuzzleVQA datasets.
  }
  \label{tab:Puzzle_acc}
  \resizebox{\textwidth}{!}{
  \begin{tabular}{lcccccc}
  \toprule
  \multirow{2}{*}{\textbf{Method}}
  & \multicolumn{3}{c}{\textbf{Qwen2.5-VL-3B-Instruct}}
  & \multicolumn{3}{c}{\textbf{InternVL3-2B-Instruct}} \\
  \cmidrule(lr){2-4}\cmidrule(lr){5-7}
  & \textbf{PuzzleVQA} & \textbf{AlgoPuzzleVQA} & \textbf{Avg}
  & \textbf{PuzzleVQA} & \textbf{AlgoPuzzleVQA} & \textbf{Avg} \\
  \midrule
  Base (zero-shot)      & 31.80 & 21.72 & 26.76 & 31.80 & 23.00 & 27.40 \\
  High-Entropy RL       & 35.00 & 24.61 & 29.81 & 32.60 & 25.22 & 28.91 \\
  GRPO                  & 43.20 & 25.44 & 34.32 & 43.20 & 26.17 & 34.69 \\
  \rowcolor[HTML]{EFEFEF} \methodname\textsubscript{G} (Ours) & 45.00 & 26.94 & 35.97~\textcolor{blue}{(+1.65)} & 45.20 & 27.22 & 36.21~\textcolor{blue}{(+1.52)} \\
  DAPO                  & 44.60 & 25.33 & 34.97 & 44.80 & 27.72 & 36.26 \\
  \rowcolor[HTML]{EFEFEF} \methodname\textsubscript{D} (Ours) & 46.80 & 26.56 & 36.68~\textcolor{blue}{(+1.71)} & 45.20 & 28.61 & 36.91~\textcolor{blue}{(+0.65)} \\
  \bottomrule
  \end{tabular}
  }
\end{table*}

\begin{table*}[t]
  \centering
  \setlength{\tabcolsep}{1.0pt}
  \setlength{\abovecaptionskip}{2pt}
  \caption{
  Comparison of scaling performance among GRPO, PAPO, and PEPO on Qwen2.5-VL-3B-Instruct trained with ViRL39K, evaluated across multiple reasoning benchmarks using the avg@8 metric. GRPO and PAPO results are taken directly from~\cite{wang2025perception}.
  }
  \label{tab:scale_perf}
  \resizebox{\textwidth}{!}{
  \begin{tabular}{lcccccccc}
  \toprule
  \textbf{Method} & \textbf{Geometry3K\textsubscript{test}} & \textbf{MathVista} & \textbf{We-Math} & \textbf{MathVerse} & \textbf{LogicVista} & \textbf{Counting} & \textbf{MMMU-Pro} & \textbf{Avg} \\
  \midrule
  GRPO  & 28.72 & 59.34 & 58.90 & 55.25 & 38.14 & 55.81 & 25.66 & 45.97 \\
  PAPO\textsubscript{G}  & 30.95 & 61.38 & 60.09 & 57.14 & 38.67 & 62.56 & 27.11 & 48.27 \\
  \rowcolor[HTML]{EFEFEF} PEPO\textsubscript{G} (Ours)  & 29.85 & 63.48 & 60.98 & 56.49 & 39.85 & 70.75 & 27.47 & 49.84~\textcolor{blue}{(+3.87)} \\
  \bottomrule
  \end{tabular}
  }
\end{table*}

\myPara{Few-shot classification.}
Following~\cite{liu2025visual}, we evaluate the training effectiveness of \methodname\ under limited data settings.
As shown in \tabref{tab:fewshot_combined}, \methodname\ consistently outperforms GRPO across the 1-shot, 2-shot, and 4-shot settings, markedly enhancing few-shot classification performance on both FGVC Aircraft and Flower102 datasets.
Compared to GRPO, \methodname\textsubscript{G} achieves average gains of +5.32 and +1.46 points on the two datasets, respectively.
These results demonstrate that our fine-grained advantage modulation enhances training effectiveness by enabling the policy model to better leverage limited supervision for improved generalization.

\myPara{Visual puzzle reasoning.}
Similar to the geometry reasoning results, \tabref{tab:Puzzle_acc} presents that \methodname~achieves consistent improvements on both in-domain and out-of-domain puzzle reasoning benchmarks.
Across the two backbones, \methodname~obtains higher accuracy than GRPO and DAPO on both PuzzleVQA and AlgoPuzzleVQA.
The relative gain is also evident on the out-of-domain AlgoPuzzleVQA dataset, where reasoning requires recognizing abstract relational and compositional patterns beyond surface visual cues.
These results indicate that the proposed visual similarity weighting enhances visuospatial reasoning and facilitates transfer to unseen puzzle configurations.

\myPara{Scalability analysis.}
To examine how \methodname\ scales with larger datasets and more complex reasoning scenarios, we train the Qwen2.5-VL-3B-Instruct model on the ViRL39K dataset and evaluate it on several reasoning benchmarks.
We follow the training configurations of PAPO~\cite{wang2025perception} in terms of learning rate and KL penalty coefficient, and adopt the same prompt and reward design to ensure a fair comparison.
As shown in Tab.~\ref{tab:scale_perf}, \methodname\textsubscript{G} achieves higher average performance than GRPO and PAPO\textsubscript{G}, demonstrating superior scalability with larger datasets and more complex reasoning tasks.
The gains are particularly evident on perception-intensive datasets such as MathVista and Counting, where visual grounding is crucial for reasoning.

\myPara{Efficiency and computational overhead.}
We evaluate the training efficiency of PEPO in Tab.~\ref{tab:overhead} by reporting end-to-end throughput (iterations per second), mean response length, and the step-level computational overhead ratio $\rho$.
The overhead ratio is defined as
$\rho={t_{\text{weight}}}/{t_{\text{step}}}$,
where $t_{\text{weight}}$ denotes the time required to compute $\mathrm{VS}_t$, $H_t$, and $w_t$, and $t_{\text{step}}$ is the total time of a reinforcement learning update step.
Across all benchmarks, $\rho$ remains below 1\%, indicating that the additional computation introduced by PEPO is negligible relative to the overall training cost.
Throughput is comparable to, and in some cases slightly higher than, that of GRPO.
We also observe that PEPO tends to produce shorter responses during training.
This reduction in average generation length partially compensates for the minor overhead of weight computation, resulting in similar or improved effective throughput.

\begin{table}[!t]
  \centering
  \setlength{\tabcolsep}{13pt}
  \setlength{\abovecaptionskip}{2pt}
  \caption{
  Training efficiency comparison between GRPO and PEPO$_G$.
  We report throughput (iterations per second), mean response length, and the step-level computational overhead ratio $\rho$ across datasets.
  The overhead ratio $\rho$ measures the proportion of additional weight computation time within a full RL update step.
  }
  \label{tab:overhead}
  \resizebox{0.95\textwidth}{!}{
  \begin{tabular}{llcccc}
    \toprule
    \textbf{Metric} & \textbf{Method}
    & \textbf{Geometry3K} & \textbf{PuzzleVQA}
    & \textbf{RefCOCO} & \textbf{FGVC Aircraft} \\
    \midrule
    \multirow{2}{*}{Iter/s}
      & GRPO     & 0.022 & 0.066 & 0.080 & 0.071 \\
      & PEPO$_G$ & 0.022 & 0.074 & 0.084 & 0.083 \\
    \midrule
    \multirow{2}{*}{Mean length}
      & GRPO     & 273 & 104 & 77 & 92 \\
      & PEPO$_G$ & 270 & 85 & 64 & 77 \\
    \midrule
    $\rho$
      & PEPO$_G$ & 0.0039 & 0.0065 & 0.0061 & 0.0048 \\
    \bottomrule
  \end{tabular}
  }
\end{table}

\begin{table}[!t]
  \centering
  \small

  \begin{minipage}[t]{0.52\linewidth}
    \centering
    \setlength{\tabcolsep}{9.5pt}
    \setlength{\abovecaptionskip}{2pt}
    \renewcommand{\arraystretch}{1.06}
    \caption{Ablation of \methodname{} components. Perception-only uses visual similarity for weighting ($\alpha=0$), while Exploration-only relies on token entropy.}
    \label{tab:ablation_methodname_qwen}
    \begin{tabular}{lcc}
      \toprule
      \textbf{Variant}
      & \textbf{Geometry3K\textsubscript{val}}
      & \textbf{PuzzleVQA}  \\
      \midrule
      GRPO (baseline)
      & 19.00
      & 43.20  \\
      Exploration-only
      & 20.18
      & 41.60  \\
      Perception-only
      & 21.07
      & 43.20  \\
      \rowcolor[HTML]{EFEFEF} \methodname\textsubscript{G}
      & 22.80
      & 45.00 \\
      \bottomrule
    \end{tabular}
  \end{minipage}\hfill
  \begin{minipage}[t]{0.46\linewidth}
    \centering
    \setlength{\tabcolsep}{9.5pt}
    \setlength{\abovecaptionskip}{2pt}
    \renewcommand{\arraystretch}{0.9}
    \caption{
    Sensitivity to the weight $\alpha$. Results are reported on FGVC (4-shot), PuzzleVQA, and geometry reasoning.
    }
    \label{tab:alpha_sweep}
      \begin{tabular}{lcc}
        \toprule
        \textbf{$\alpha$} & \textbf{Geometry (Avg)} & \textbf{FGVC (4-shot)} \\
        \midrule
        GRPO & 32.64 & 63.94 \\
        0.00 & 36.17 & 75.48 \\
        0.02 & 36.01 & 75.79 \\
        0.05 & 36.70 & 72.97 \\
        0.10 & 36.31 & 68.17 \\
        0.15 & 36.02 & 70.39 \\
        \bottomrule
      \end{tabular}%
  \end{minipage}
\end{table}

\subsection{Ablation Study}

We conduct ablation experiments on the Qwen2.5-VL-3B-Instruct model to examine the effectiveness of \methodname.
The analysis includes the effect of visual and entropy components and the influence of different weighting schemes.

\myPara{Component analysis.}
\tabref{tab:ablation_methodname_qwen} analyzes the effect of individual components in \methodname~on Geometry3K, PuzzleVQA, and RefCOCO.
Using only visual similarity (perception-only) enhances perceptual grounding but limits reasoning diversity, while relying solely on entropy (exploration-only) introduces unstable optimization and reduced accuracy.
The full visual-entropy formulation integrates both factors and achieves the best results, indicating their complementary roles in stabilizing learning and encouraging diverse yet coherent reasoning.
These results show that balanced perception-exploration coupling is essential for token-level policy optimization in multimodal reasoning.

\myPara{Sensitivity and robustness of $\alpha$.}
Tab.~\ref{tab:alpha_sweep} analyzes the impact of the weighting coefficient $\alpha$ on both perception-intensive (FGVC Aircraft) and reasoning-oriented (Geometry reasoning) benchmarks.
All non-zero $\alpha$ settings consistently outperform the GRPO baseline across tasks.
For Geometry reasoning, introducing a small positive $\alpha$ leads to clear improvements, and performance remains stable across a moderate range of values, indicating limited sensitivity to precise tuning.
For FGVC Aircraft, smaller $\alpha$ values yield the strongest gains, while larger values gradually reduce the improvement, although performance remains above the baseline throughout.

\myPara{Weighting design.}
Tab.~\ref{tab:weight_ablation} evaluates the effect of different weighting formulations on Geometry3K.
Removing the gradual scheduling strategy leads to a substantial performance drop, indicating that progressive modulation of token weights is important for stable optimization.
Excluding per-sample min-max normalization also degrades performance, suggesting that normalization helps maintain consistent scaling across tokens within each response.
Replacing the gated formulation with a simple additive fusion further reduces accuracy.
This result indicates that the bounded and mean-centered gating mechanism provides a more stable integration of perceptual and entropy signals than direct summation.

\myPara{Layer selection.}
Tab.~\ref{tab:layer_ablation} analyzes the impact of restricting the hidden-state layers used for weight computation.
Using features from shallow, intermediate, or deep layers alone consistently underperforms the default configuration.
In contrast, aggregating signals across all layers yields the highest accuracy, suggesting that visual relevance is distributed across the model hierarchy rather than localized to a specific depth.

\begin{table}[!t]
  \centering
  \scriptsize
  \setlength{\tabcolsep}{6pt}

  \begin{minipage}[t]{0.48\linewidth}
  \centering
  \setlength{\tabcolsep}{12pt}
  \caption{
  Removing scheduling, normalization, or gating degrades performance.
  }
  \label{tab:weight_ablation}
  \begin{tabular}{lcc}
  \toprule
  \textbf{Variant}
  & \textbf{Geometry3K\textsubscript{val}}
  & \textbf{Geometry3K\textsubscript{Test}} \\
  \midrule
  \rowcolor[HTML]{EFEFEF} \methodname\textsubscript{G}
  & 22.80 & 26.81 \\
  W/o schedule
  & 19.80 & 24.54 \\
  W/o min-max
  & 21.91 & 26.93 \\
  Additive fusion
  & 20.99 & 22.94 \\
  \bottomrule
  \end{tabular}
  \end{minipage}
  \hfill
  \begin{minipage}[t]{0.50\linewidth}
  \centering
  \setlength{\tabcolsep}{14.0pt}
  \renewcommand{\arraystretch}{1.23}
  \caption{
  Effect of restricting the hidden-state layers used for weight computation in \methodname.
  }
  \label{tab:layer_ablation}
  \begin{tabular}{lcc}
  \toprule
  \textbf{Layer range}
  & \textbf{Geometry3K\textsubscript{val}}
  & \textbf{Geometry3K\textsubscript{Test}} \\
  \midrule
  1--10
  & 18.92 & 23.71 \\
  11--20
  & 22.04 & 25.75 \\
  21--32
  & 19.89 & 24.94 \\
  \bottomrule
  \end{tabular}
  \end{minipage}
\end{table}

\begin{figure*}[t]
  \centering
  \includegraphics[width=\textwidth]{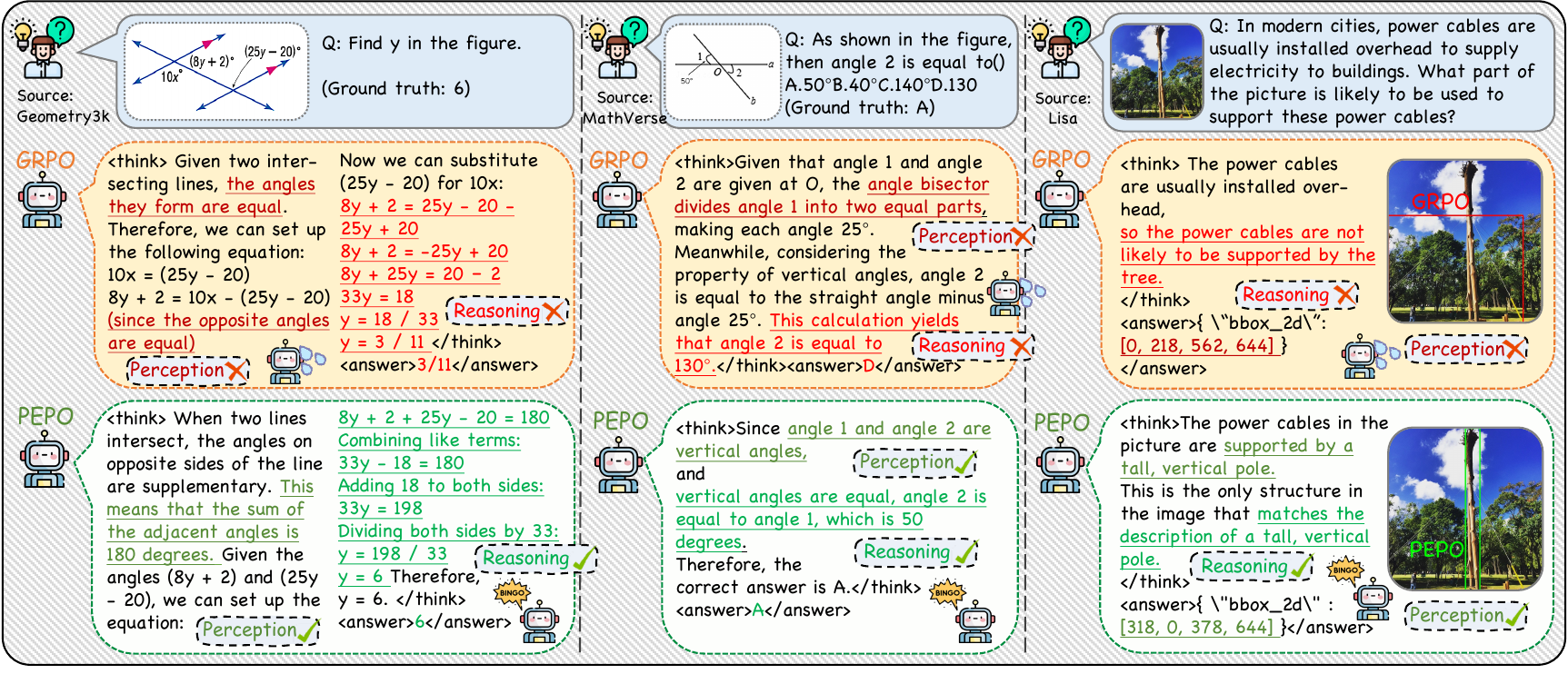}
  \caption{
        Qualitative comparison on Geometry3K, MathVerse, and LISA datasets.
        The GRPO-trained model exhibits perception failures and inconsistent reasoning, leading to incorrect answers.
        In contrast, the PEPO-trained model generates coherent, visually grounded reasoning chains that produce correct results, demonstrating the effectiveness of PEPO in enhancing multimodal reasoning.
    }
  \label{fig:case}
\end{figure*}

\subsection{Qualitative Comparisons}
\label{sec:qualitative}

\myPara{Case studies.}
Fig.~\ref{fig:case} presents representative examples from geometry reasoning (Geometry3K), mathematical derivation (MathVerse), and visual grounding (LISA).
In the geometry example, GRPO incorrectly manipulates angle relations despite the diagram indicating supplementary structure, revealing a failure to align algebraic operations with visual constraints.
In the mathematical derivation case, GRPO introduces inconsistent intermediate conclusions, leading to a contradiction in the final answer.
For visual grounding, GRPO misidentifies salient regions in the image, reflecting insufficient reliance on perceptual evidence.
In contrast, \methodname\ maintains alignment between visual cues and intermediate reasoning steps.
It correctly extracts geometric relations from the diagram, preserves logical consistency in algebraic transformations, and grounds its predictions on visually relevant structures in the image.
Across tasks, the reasoning trajectories generated by \methodname\ remain coherent and consistent with the available visual evidence, resulting in correct final predictions.

\begin{figure}[t]
  \centering
  \includegraphics[width=\linewidth]{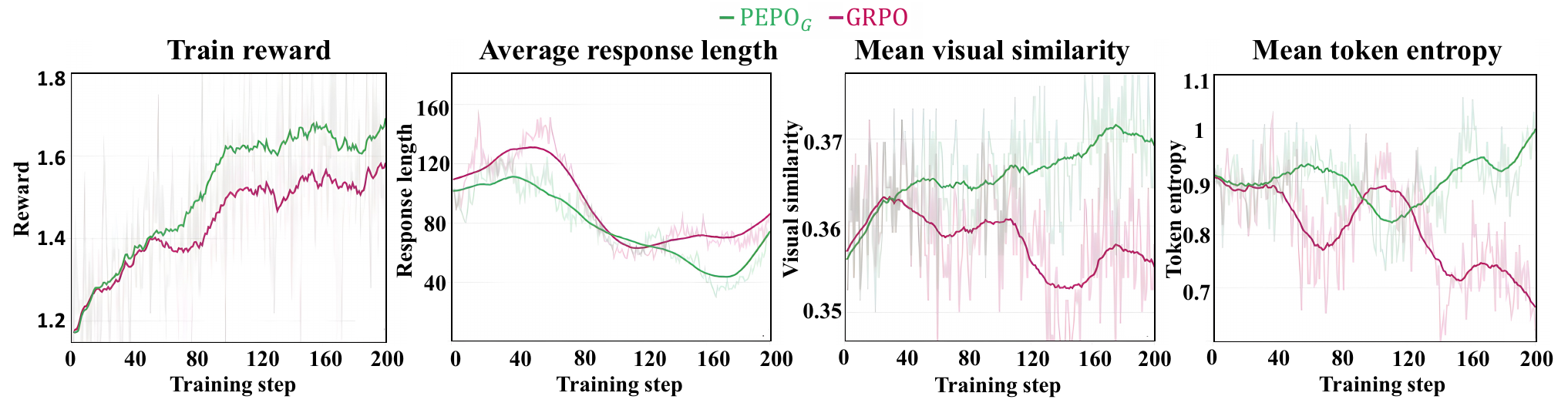}
  \caption{
  Training curves on FGVC Aircraft (4-shot).
  We compare GRPO and PEPO$_G$ in terms of training reward, mean response length, mean visual similarity, and mean entropy.
  }
  \label{fig:fgvc_curve}
\end{figure}

\myPara{Training dynamics.}
Fig.~\ref{fig:fgvc_curve} compares the optimization behavior of GRPO and PEPO$_G$ on FGVC Aircraft (4-shot).
PEPO$_G$ attains consistently higher training reward and converges to a stronger plateau.
The mean response length under PEPO$_G$ decreases more gradually during training, resulting in shorter generations.
In addition, PEPO$_G$ exhibits a progressively increasing mean visual similarity, whereas GRPO shows a weaker or declining trend.
Regarding entropy, GRPO undergoes a sharper decay, while PEPO$_G$ maintains a more moderate entropy level throughout training.
These results indicate that PEPO$_G$ yields more stable optimization dynamics across reward, generation length, visual alignment, and exploration behavior.

\section{Conclusions}

We present {Perception-Exploration Policy Optimization} (\methodname), a token-level reinforcement learning framework for large vision-language models.
By coupling visual similarity and token entropy through a smooth gating mechanism, \methodname~achieves fine-grained advantage estimation that jointly models perceptual grounding and reasoning exploration.
Based on two different architectures, extensive experiments demonstrate that \methodname~consistently outperforms GRPO and DAPO across geometry reasoning, visual puzzles, visual grounding, and few-shot classification, while maintaining training stability and scalability on large multimodal datasets.
These results confirm that integrating perception and exploration offers a principled and effective approach to advancing multimodal reasoning in LVLMs.

\bibliography{main}

@String(AAAI  = {AAAI})

@article{wang2025perception,
  title={Perception-aware policy optimization for multimodal reasoning},
  author={Wang, Zhenhailong and Guo, Xuehang and Stoica, Sofia and Xu, Haiyang and Wang, Hongru and Ha, Hyeonjeong and Chen, Xiusi and Chen, Yangyi and Yan, Ming and Huang, Fei and others},
  journal={arXiv preprint arXiv:2507.06448},
  year={2025}
}

@article{huang2025spotlight,
  title={Spotlight on Token Perception for Multimodal Reinforcement Learning},
  author={Huang, Siyuan and Qu, Xiaoye and Li, Yafu and Luo, Yun and He, Zefeng and Liu, Daizong and Cheng, Yu},
  journal={arXiv preprint arXiv:2510.09285},
  year={2025}
}

@inproceedings{jian2025look,
  title={Look again, think slowly: Enhancing visual reflection in vision-language models},
  author={Jian, Pu and Wu, Junhong and Sun, Wei and Wang, Chen and Ren, Shuo and Zhang, Jiajun},
  booktitle={Proceedings of the 2025 Conference on Empirical Methods in Natural Language Processing},
  pages={9262--9281},
  year={2025}
}

@article{schulman2017proximal,
  title={Proximal Policy Optimization Algorithms},
  author={Schulman, John and Wolski, Filip and Dhariwal, Prafulla and Radford, Alec and Klimov, Oleg},
  journal={arXiv preprint arXiv:1707.06347},
  year={2017}
}

@article{wei2022chain,
  title={Chain-of-thought prompting elicits reasoning in large language models},
  author={Wei, Jason and Wang, Xuezhi and Schuurmans, Dale and Bosma, Maarten and Xia, Fei and Chi, Ed and Le, Quoc V and Zhou, Denny and others},
  journal={Advances in neural information processing systems},
  volume={35},
  pages={24824--24837},
  year={2022}
}

@article{shao2024deepseekmath,
  title={Deepseekmath: Pushing the limits of mathematical reasoning in open language models},
  author={Shao, Zhihong and Wang, Peiyi and Zhu, Qihao and Xu, Runxin and Song, Junxiao and Bi, Xiao and Zhang, Haowei and Zhang, Mingchuan and Li, YK and Wu, Yang and others},
  journal={arXiv preprint arXiv:2402.03300},
  year={2024}
}

@article{guo2025deepseek,
  title={Deepseek-r1 incentivizes reasoning in llms through reinforcement learning},
  author={Guo, Daya and Yang, Dejian and Zhang, Haowei and Song, Junxiao and Wang, Peiyi and Zhu, Qihao and Xu, Runxin and Zhang, Ruoyu and Ma, Shirong and Bi, Xiao and others},
  journal={Nature},
  volume={645},
  number={8081},
  pages={633--638},
  year={2025},
  publisher={Nature Publishing Group UK London}
}

@article{wang2025beyond,
  title={Beyond the 80/20 rule: High-entropy minority tokens drive effective reinforcement learning for llm reasoning},
  author={Wang, Shenzhi and Yu, Le and Gao, Chang and Zheng, Chujie and Liu, Shixuan and Lu, Rui and Dang, Kai and Chen, Xionghui and Yang, Jianxin and Zhang, Zhenru and others},
  journal={arXiv preprint arXiv:2506.01939},
  year={2025}
}

@article{wang2025harnessing,
  title={Harnessing uncertainty: Entropy-modulated policy gradients for long-horizon llm agents},
  author={Wang, Jiawei and Liu, Jiacai and Fu, Yuqian and Li, Yingru and Wang, Xintao and Lin, Yuan and Yue, Yu and Zhang, Lin and Wang, Yang and Wang, Ke},
  journal={arXiv preprint arXiv:2509.09265},
  year={2025}
}

@article{yu2025dapo,
  title={Dapo: An open-source llm reinforcement learning system at scale},
  author={Yu, Qiying and Zhang, Zheng and Zhu, Ruofei and Yuan, Yufeng and Zuo, Xiaochen and Yue, Yu and Dai, Weinan and Fan, Tiantian and Liu, Gaohong and Liu, Lingjun and others},
  journal={arXiv preprint arXiv:2503.14476},
  year={2025}
}

@article{cui2025entropy,
  title={The entropy mechanism of reinforcement learning for reasoning language models},
  author={Cui, Ganqu and Zhang, Yuchen and Chen, Jiacheng and Yuan, Lifan and Wang, Zhi and Zuo, Yuxin and Li, Haozhan and Fan, Yuchen and Chen, Huayu and Chen, Weize and others},
  journal={arXiv preprint arXiv:2505.22617},
  year={2025}
}

@article{vanlioglu2025entropy,
  title={Entropy-guided sequence weighting for efficient exploration in RL-based LLM fine-tuning},
  author={Vanlioglu, Abdullah},
  journal={arXiv preprint arXiv:2503.22456},
  year={2025}
}

@article{chen2025seed,
  title={Seed-grpo: Semantic entropy enhanced grpo for uncertainty-aware policy optimization},
  author={Chen, Minghan and Chen, Guikun and Wang, Wenguan and Yang, Yi},
  journal={arXiv preprint arXiv:2505.12346},
  year={2025}
}

@article{liang2025modomodo,
  title={MoDoMoDo: Multi-Domain Data Mixtures for Multimodal LLM Reinforcement Learning},
  author={Liang, Yiqing and Qiu, Jielin and Ding, Wenhao and Liu, Zuxin and Tompkin, James and Xu, Mengdi and Xia, Mengzhou and Tu, Zhengzhong and Shi, Laixi and Zhu, Jiacheng},
  journal={arXiv preprint arXiv:2505.24871},
  year={2025}
}

@misc{chen2025vinci,
  title={Vinci. R1-v: Reinforcing super generalization ability in vision-language models with less than 3},
  author={Chen, Liang and Li, Lei and Zhao, Haozhe and Song, Yifan},
  journal={arXiv preprint arXiv:2308.15363},
  year={2025}
}

@article{liu2025noisyrollout,
  title={Noisyrollout: Reinforcing visual reasoning with data augmentation},
  author={Liu, Xiangyan and Ni, Jinjie and Wu, Zijian and Du, Chao and Dou, Longxu and Wang, Haonan and Pang, Tianyu and Shieh, Michael Qizhe},
  journal={arXiv preprint arXiv:2504.13055},
  year={2025}
}

@article{yang2025r1,
  title={R1-Onevision: Advancing Generalized Multimodal Reasoning through Cross-Modal Formalization},
  author={Yang, Yi and He, Xiaoxuan and Pan, Hongkun and Jiang, Xiyan and Deng, Yan and Yang, Xingtao and Lu, Haoyu and Yin, Dacheng and Rao, Fengyun and Zhu, Minfeng and others},
  journal={arXiv preprint arXiv:2503.10615},
  year={2025}
}

@article{meng2025mm,
  title={MM-Eureka: Exploring the Frontiers of Multimodal Reasoning with Rule-based Reinforcement Learning},
  author={Meng, Fanqing and Du, Lingxiao and Liu, Zongkai and Zhou, Zhixiang and Lu, Quanfeng and Fu, Daocheng and Han, Tiancheng and Shi, Botian and Wang, Wenhai and He, Junjun and others},
  journal={arXiv preprint arXiv:2503.07365},
  year={2025}
}

@article{wang2025internvl3.5,
  title={InternVL3. 5: Advancing Open-Source Multimodal Models in Versatility, Reasoning, and Efficiency},
  author={Wang, Weiyun and Gao, Zhangwei and Gu, Lixin and Pu, Hengjun and Cui, Long and Wei, Xingguang and Liu, Zhaoyang and Jing, Linglin and Ye, Shenglong and Shao, Jie and others},
  journal={arXiv preprint arXiv:2508.18265},
  year={2025}
}

@article{yuan2025vl,
  title={VL-Cogito: Progressive Curriculum Reinforcement Learning for Advanced Multimodal Reasoning},
  author={Yuan, Ruifeng and Xiao, Chenghao and Leng, Sicong and Wang, Jianyu and Li, Long and Xu, Weiwen and Chan, Hou Pong and Zhao, Deli and Xu, Tingyang and Wei, Zhongyu and others},
  journal={arXiv preprint arXiv:2507.22607},
  year={2025}
}

@article{bai2025univg,
  title={UniVG-R1: Reasoning Guided Universal Visual Grounding with Reinforcement Learning},
  author={Bai, Sule and Li, Mingxing and Liu, Yong and Tang, Jing and Zhang, Haoji and Sun, Lei and Chu, Xiangxiang and Tang, Yansong},
  journal={arXiv preprint arXiv:2505.14231},
  year={2025}
}

@article{wang2025sota,
  title={SoTA with Less: MCTS-Guided Sample Selection for Data-Efficient Visual Reasoning Self-Improvement},
  author={Wang, Xiyao and Yang, Zhengyuan and Feng, Chao and Lu, Hongjin and Li, Linjie and Lin, Chung-Ching and Lin, Kevin and Huang, Furong and Wang, Lijuan},
  journal={arXiv preprint arXiv:2504.07934},
  year={2025}
}

@article{zhu2025shuffle,
  title={Shuffle-R1: Efficient RL framework for Multimodal Large Language Models via Data-centric Dynamic Shuffle},
  author={Zhu, Linghao and Guan, Yiran and Liang, Dingkang and Ju, Jianzhong and Luo, Zhenbo and Qin, Bin and Luan, Jian and Liu, Yuliang and Bai, Xiang},
  journal={arXiv preprint arXiv:2508.05612},
  year={2025}
}

@article{qiao2025we2,
  title={We-Math 2.0: A Versatile MathBook System for Incentivizing Visual Mathematical Reasoning},
  author={Qiao, Runqi and Tan, Qiuna and Yang, Peiqing and Wang, Yanzi and Wang, Xiaowan and Wan, Enhui and Zhou, Sitong and Dong, Guanting and Zeng, Yuchen and Xu, Yida and others},
  journal={arXiv preprint arXiv:2508.10433},
  year={2025}
}

@article{wang2025vicrit,
  title={ViCrit: A Verifiable Reinforcement Learning Proxy Task for Visual Perception in VLMs},
  author={Wang, Xiyao and Yang, Zhengyuan and Feng, Chao and Liang, Yongyuan and Zhou, Yuhang and Liu, Xiaoyu and Zang, Ziyi and Li, Ming and Lin, Chung-Ching and Lin, Kevin and others},
  journal={arXiv preprint arXiv:2506.10128},
  year={2025}
}

@article{deng2025openvlthinker,
  title={Openvlthinker: An early exploration to complex vision-language reasoning via iterative self-improvement},
  author={Deng, Yihe and Bansal, Hritik and Yin, Fan and Peng, Nanyun and Wang, Wei and Chang, Kai-Wei},
  journal={arXiv preprint arXiv:2503.17352},
  year={2025}
}

@article{ai2025m2,
  title={M2-reasoning: Empowering mllms with unified general and spatial reasoning},
  author={AI, Inclusion and Wang, Fudong and Liu, Jiajia and Chen, Jingdong and Zhou, Jun and Ji, Kaixiang and Ru, Lixiang and Guo, Qingpei and Zheng, Ruobing and Li, Tianqi and others},
  journal={arXiv preprint arXiv:2507.08306},
  year={2025}
}

@article{chen2025g1,
  title={G1: Bootstrapping Perception and Reasoning Abilities of Vision-Language Model via Reinforcement Learning},
  author={Chen, Liang and Gao, Hongcheng and Liu, Tianyu and Huang, Zhiqi and Sung, Flood and Zhou, Xinyu and Wu, Yuxin and Chang, Baobao},
  journal={arXiv preprint arXiv:2505.13426},
  year={2025}
}

@article{wang2025vl,
  title={Vl-rethinker: Incentivizing self-reflection of vision-language models with reinforcement learning},
  author={Wang, Haozhe and Qu, Chao and Huang, Zuming and Chu, Wei and Lin, Fangzhen and Chen, Wenhu},
  journal={arXiv preprint arXiv:2504.08837},
  year={2025}
}

@article{wan2025srpo,
  title={Srpo: Enhancing multimodal llm reasoning via reflection-aware reinforcement learning},
  author={Wan, Zhongwei and Dou, Zhihao and Liu, Che and Zhang, Yu and Cui, Dongfei and Zhao, Qinjian and Shen, Hui and Xiong, Jing and Xin, Yi and Jiang, Yifan and others},
  journal={arXiv preprint arXiv:2506.01713},
  year={2025}
}

@article{shen2025vlm,
  title={VLM-R1: A Stable and Generalizable R1-style Large Vision-Language Model},
  author={Shen, Haozhan and Liu, Peng and Li, Jingcheng and Fang, Chunxin and Ma, Yibo and Liao, Jiajia and Shen, Qiaoli and Zhang, Zilun and Zhao, Kangjia and Zhang, Qianqian and others},
  journal={arXiv preprint arXiv:2504.07615},
  year={2025}
}

@article{liu2025visual,
  title={Visual-RFT: Visual Reinforcement Fine-Tuning},
  author={Liu, Ziyu and Sun, Zeyi and Zang, Yuhang and Dong, Xiaoyi and Cao, Yuhang and Duan, Haodong and Lin, Dahua and Wang, Jiaqi},
  journal={arXiv preprint arXiv:2503.01785},
  year={2025}
}

@article{tan2025reason,
  title={Reason-rft: Reinforcement fine-tuning for visual reasoning},
  author={Tan, Huajie and Ji, Yuheng and Hao, Xiaoshuai and Lin, Minglan and Wang, Pengwei and Wang, Zhongyuan and Zhang, Shanghang},
  journal={arXiv preprint arXiv:2503.20752},
  year={2025}
}

@article{li2025think,
  title={Think or not think: A study of explicit thinking in rule-based visual reinforcement fine-tuning},
  author={Li, Ming and Zhong, Jike and Zhao, Shitian and Lai, Yuxiang and Zhang, Haoquan and Zhu, Wang Bill and Zhang, Kaipeng},
  journal={arXiv preprint arXiv:2503.16188},
  year={2025}
}

@article{shen2025satori,
  title={Satori-r1: Incentivizing multimodal reasoning with spatial grounding and verifiable rewards},
  author={Shen, Chuming and Wei, Wei and Qu, Xiaoye and Cheng, Yu},
  journal={arXiv preprint arXiv:2505.19094},
  year={2025}
}

@inproceedings{yu2025docthinker,
  title={Docthinker: Explainable multimodal large language models with rule-based reinforcement learning for document understanding},
  author={Yu, Wenwen and Yang, Zhibo and Liu, Yuliang and Bai, Xiang},
  booktitle={Proceedings of the IEEE/CVF International Conference on Computer Vision},
  pages={837--847},
  year={2025}
}

@article{gou2025perceptual,
  title={Perceptual Decoupling for Scalable Multi-modal Reasoning via Reward-Optimized Captioning},
  author={Gou, Yunhao and Chen, Kai and Liu, Zhili and Hong, Lanqing and Jin, Xin and Li, Zhenguo and Kwok, James T and Zhang, Yu},
  journal={arXiv preprint arXiv:2506.04559},
  year={2025}
}

@article{yu2025perception,
  title={Perception-r1: Pioneering perception policy with reinforcement learning},
  author={Yu, En and Lin, Kangheng and Zhao, Liang and Yin, Jisheng and Wei, Yana and Peng, Yuang and Wei, Haoran and Sun, Jianjian and Han, Chunrui and Ge, Zheng and others},
  journal={arXiv preprint arXiv:2504.07954},
  year={2025}
}

@article{wu2025visualquality,
  title={VisualQuality-R1: Reasoning-Induced Image Quality Assessment via Reinforcement Learning to Rank},
  author={Wu, Tianhe and Zou, Jian and Liang, Jie and Zhang, Lei and Ma, Kede},
  journal={arXiv preprint arXiv:2505.14460},
  year={2025}
}

@article{liu2025visionreasoner,
  title={VisionReasoner: Unified Visual Perception and Reasoning via Reinforcement Learning},
  author={Liu, Yuqi and Qu, Tianyuan and Zhong, Zhisheng and Peng, Bohao and Liu, Shu and Yu, Bei and Jia, Jiaya},
  journal={arXiv preprint arXiv:2505.12081},
  year={2025}
}

@article{jiang2025rex,
  title={Rex-Thinker: Grounded Object Referring via Chain-of-Thought Reasoning},
  author={Jiang, Qing and Chen, Xingyu and Zeng, Zhaoyang and Yu, Junzhi and Zhang, Lei},
  journal={arXiv preprint arXiv:2506.04034},
  year={2025}
}

@article{ni2025point,
  title={Point-rft: Improving multimodal reasoning with visually grounded reinforcement finetuning},
  author={Ni, Minheng and Yang, Zhengyuan and Li, Linjie and Lin, Chung-Ching and Lin, Kevin and Zuo, Wangmeng and Wang, Lijuan},
  journal={arXiv preprint arXiv:2505.19702},
  year={2025}
}

@article{fan2025sophiavl,
  title={SophiaVL-R1: Reinforcing MLLMs Reasoning with Thinking Reward},
  author={Fan, Kaixuan and Feng, Kaituo and Lyu, Haoming and Zhou, Dongzhan and Yue, Xiangyu},
  journal={arXiv preprint arXiv:2505.17018},
  year={2025}
}

@article{ma2025one,
  title={One RL to See Them All: Visual Triple Unified Reinforcement Learning},
  author={Ma, Yan and Du, Linge and Shen, Xuyang and Chen, Shaoxiang and Li, Pengfei and Ren, Qibing and Ma, Lizhuang and Dai, Yuchao and Liu, Pengfei and Yan, Junjie},
  journal={arXiv preprint arXiv:2505.18129},
  year={2025}
}

@article{chen2025grpo,
  title={GRPO-CARE: Consistency-Aware Reinforcement Learning for Multimodal Reasoning},
  author={Chen, Yi and Ge, Yuying and Wang, Rui and Ge, Yixiao and Cheng, Junhao and Shan, Ying and Liu, Xihui},
  journal={arXiv preprint arXiv:2506.16141},
  year={2025}
}

@article{su2025pixel,
  title={Pixel reasoner: Incentivizing pixel-space reasoning with curiosity-driven reinforcement learning},
  author={Su, Alex and Wang, Haozhe and Ren, Weiming and Lin, Fangzhen and Chen, Wenhu},
  journal={arXiv preprint arXiv:2505.15966},
  year={2025}
}

@article{xiao2025advancing,
  title={Advancing Multimodal Reasoning Capabilities of Multimodal Large Language Models via Visual Perception Reward},
  author={Xiao, Tong and Xu, Xin and Huang, Zhenya and Gao, Hongyu and Liu, Quan and Liu, Qi and Chen, Enhong},
  journal={arXiv preprint arXiv:2506.07218},
  year={2025}
}

@article{wu2025reinforcing,
  title={Reinforcing spatial reasoning in vision-language models with interwoven thinking and visual drawing},
  author={Wu, Junfei and Guan, Jian and Feng, Kaituo and Liu, Qiang and Wu, Shu and Wang, Liang and Wu, Wei and Tan, Tieniu},
  journal={arXiv preprint arXiv:2506.09965},
  year={2025}
}

@article{zhang2025chain,
  title={Chain-of-Focus: Adaptive Visual Search and Zooming for Multimodal Reasoning via RL},
  author={Zhang, Xintong and Gao, Zhi and Zhang, Bofei and Li, Pengxiang and Zhang, Xiaowen and Liu, Yang and Yuan, Tao and Wu, Yuwei and Jia, Yunde and Zhu, Song-Chun and others},
  journal={arXiv preprint arXiv:2505.15436},
  year={2025}
}

@article{jiang2025vlm,
  title={VLM-R 3: Region Recognition, Reasoning, and Refinement for Enhanced Multimodal Chain-of-Thought},
  author={Jiang, Chaoya and Heng, Yongrui and Ye, Wei and Yang, Han and Xu, Haiyang and Yan, Ming and Zhang, Ji and Huang, Fei and Zhang, Shikun},
  journal={arXiv preprint arXiv:2505.16192},
  year={2025}
}

@article{sarch2025grounded,
  title={Grounded Reinforcement Learning for Visual Reasoning},
  author={Sarch, Gabriel and Saha, Snigdha and Khandelwal, Naitik and Jain, Ayush and Tarr, Michael J and Kumar, Aviral and Fragkiadaki, Katerina},
  journal={arXiv preprint arXiv:2505.23678},
  year={2025}
}

@article{li2025relation,
  title={Relation-R1: Progressively Cognitive Chain-of-Thought Guided Reinforcement Learning for Unified Relation Comprehension},
  author={Li, Lin and Chen, Wei and Li, Jiahui and Cheng, Kwang-Ting and Chen, Long},
  journal={arXiv preprint arXiv:2504.14642},
  year={2025}
}

@article{xu2025visual,
  title={Visual Planning: Let's Think Only with Images},
  author={Xu, Yi and Li, Chengzu and Zhou, Han and Wan, Xingchen and Zhang, Caiqi and Korhonen, Anna and Vuli{'c}, Ivan},
  journal={arXiv preprint arXiv:2505.11409},
  year={2025}
}

@article{su2025openthinkimg,
  title={Openthinkimg: Learning to think with images via visual tool reinforcement learning},
  author={Su, Zhaochen and Li, Linjie and Song, Mingyang and Hao, Yunzhuo and Yang, Zhengyuan and Zhang, Jun and Chen, Guanjie and Gu, Jiawei and Li, Juntao and Qu, Xiaoye and others},
  journal={arXiv preprint arXiv:2505.08617},
  year={2025}
}

@article{zheng2025deepeyes,
  title={DeepEyes: Incentivizing" Thinking with Images" via Reinforcement Learning},
  author={Zheng, Ziwei and Yang, Michael and Hong, Jack and Zhao, Chenxiao and Xu, Guohai and Yang, Le and Shen, Chao and Yu, Xing},
  journal={arXiv preprint arXiv:2505.14362},
  year={2025}
}

@article{fan2025grit,
  title={GRIT: Teaching MLLMs to Think with Images},
  author={Fan, Yue and He, Xuehai and Yang, Diji and Zheng, Kaizhi and Kuo, Ching-Chen and Zheng, Yuting and Narayanaraju, Sravana Jyothi and Guan, Xinze and Wang, Xin Eric},
  journal={arXiv preprint arXiv:2505.15879},
  year={2025}
}

@article{zhu2025active,
  title={Active-O3: Empowering Multimodal Large Language Models with Active Perception via GRPO},
  author={Zhu, Muzhi and Zhong, Hao and Zhao, Canyu and Du, Zongze and Huang, Zheng and Liu, Mingyu and Chen, Hao and Zou, Cheng and Chen, Jingdong and Yang, Ming and others},
  journal={arXiv preprint arXiv:2505.21457},
  year={2025}
}

@article{yang2025kwai,
  title={Kwai keye-vl 1.5 technical report},
  author={Yang, Biao and Wen, Bin and Ding, Boyang and Liu, Changyi and Chu, Chenglong and Song, Chengru and Rao, Chongling and Yi, Chuan and Li, Da and Zang, Dunju and others},
  journal={arXiv preprint arXiv:2509.01563},
  year={2025}
}

@article{bai2025qwen2,
  title={Qwen2. 5-vl technical report},
  author={Bai, Shuai and Chen, Keqin and Liu, Xuejing and Wang, Jialin and Ge, Wenbin and Song, Sibo and Dang, Kai and Wang, Peng and Wang, Shijie and Tang, Jun and others},
  journal={arXiv preprint arXiv:2502.13923},
  year={2025}
}

@article{zhu2025internvl3,
  title={Internvl3: Exploring advanced training and test-time recipes for open-source multimodal models},
  author={Zhu, Jinguo and Wang, Weiyun and Chen, Zhe and Liu, Zhaoyang and Ye, Shenglong and Gu, Lixin and Tian, Hao and Duan, Yuchen and Su, Weijie and Shao, Jie and others},
  journal={arXiv preprint arXiv:2504.10479},
  year={2025}
}

@article{hurst2024gpt,
  title={Gpt-4o system card},
  author={Hurst, Aaron and Lerer, Adam and Goucher, Adam P and Perelman, Adam and Ramesh, Aditya and Clark, Aidan and Ostrow, AJ and Welihinda, Akila and Hayes, Alan and Radford, Alec and others},
  journal={arXiv preprint arXiv:2410.21276},
  year={2024}
}

@article{team2024gemini,
  title={Gemini 1.5: Unlocking multimodal understanding across millions of tokens of context},
  author={Team, Gemini and Georgiev, Petko and Lei, Ving Ian and Burnell, Ryan and Bai, Libin and Gulati, Anmol and Tanzer, Garrett and Vincent, Damien and Pan, Zhufeng and Wang, Shibo and others},
  journal={arXiv preprint arXiv:2403.05530},
  year={2024}
}

@article{lu2023mathvista,
  title={Mathvista: Evaluating mathematical reasoning of foundation models in visual contexts},
  author={Lu, Pan and Bansal, Hritik and Xia, Tony and Liu, Jiacheng and Li, Chunyuan and Hajishirzi, Hannaneh and Cheng, Hao and Chang, Kai-Wei and Galley, Michel and Gao, Jianfeng},
  journal={arXiv preprint arXiv:2310.02255},
  year={2023}
}

@inproceedings{zhang2024mathverse,
  title={Mathverse: Does your multi-modal llm truly see the diagrams in visual math problems?},
  author={Zhang, Renrui and Jiang, Dongzhi and Zhang, Yichi and Lin, Haokun and Guo, Ziyu and Qiu, Pengshuo and Zhou, Aojun and Lu, Pan and Chang, Kai-Wei and Qiao, Yu and others},
  booktitle={European Conference on Computer Vision},
  pages={169--186},
  year={2024},
  organization={Springer}
}

@article{qiao2024we,
  title={We-math: Does your large multimodal model achieve human-like mathematical reasoning?},
  author={Qiao, Runqi and Tan, Qiuna and Dong, Guanting and Wu, Minhui and Sun, Chong and Song, Xiaoshuai and GongQue, Zhuoma and Lei, Shanglin and Wei, Zhe and Zhang, Miaoxuan and others},
  journal={arXiv preprint arXiv:2407.01284},
  year={2024}
}

@article{xiao2024logicvista,
  title={Logicvista: Multimodal llm logical reasoning benchmark in visual contexts},
  author={Xiao, Yijia and Sun, Edward and Liu, Tianyu and Wang, Wei},
  journal={arXiv preprint arXiv:2407.04973},
  year={2024}
}

@article{lu2021inter,
  title={Inter-gps: Interpretable geometry problem solving with formal language and symbolic reasoning},
  author={Lu, Pan and Gong, Ran and Jiang, Shibiao and Qiu, Liang and Huang, Siyuan and Liang, Xiaodan and Zhu, Song-Chun},
  journal={arXiv preprint arXiv:2105.04165},
  year={2021}
}

@inproceedings{kazemzadeh2014referitgame,
  title={Referitgame: Referring to objects in photographs of natural scenes},
  author={Kazemzadeh, Sahar and Ordonez, Vicente and Matten, Mark and Berg, Tamara},
  booktitle={Proceedings of the 2014 conference on empirical methods in natural language processing (EMNLP)},
  pages={787--798},
  year={2014}
}

@inproceedings{lai2024lisa,
  title={Lisa: Reasoning segmentation via large language model},
  author={Lai, Xin and Tian, Zhuotao and Chen, Yukang and Li, Yanwei and Yuan, Yuhui and Liu, Shu and Jia, Jiaya},
  booktitle={Proceedings of the IEEE/CVF Conference on Computer Vision and Pattern Recognition},
  pages={9579--9589},
  year={2024}
}

@inproceedings{nilsback2008flower102,
  title={Automated flower classification over a large number of classes},
  author={Nilsback, Maria-Elena and Zisserman, Andrew},
  booktitle={2008 Sixth Indian conference on computer vision, graphics image processing},
  pages={722--729},
  year={2008},
  organization={IEEE}
}

@article{maji2013fgvc,
  title={Fine-grained visual classification of aircraft},
  author={Maji, Subhransu and Rahtu, Esa and Kannala, Juho and Blaschko, Matthew and Vedaldi, Andrea},
  journal={arXiv preprint arXiv:1306.5151},
  year={2013}
}

@article{chia2024puzzlevqa,
  title={Puzzlevqa: Diagnosing multimodal reasoning challenges of language models with abstract visual patterns},
  author={Chia, Yew Ken and Han, Vernon Toh Yan and Ghosal, Deepanway and Bing, Lidong and Poria, Soujanya},
  journal={arXiv preprint arXiv:2403.13315},
  year={2024}
}

@inproceedings{yu2016modeling,
  title={Modeling context in referring expressions},
  author={Yu, Licheng and Poirson, Patrick and Yang, Shan and Berg, Alexander C and Berg, Tamara L},
  booktitle={European conference on computer vision},
  pages={69--85},
  year={2016},
  organization={Springer}
}

@inproceedings{antol2015vqa,
  title={Vqa: Visual question answering},
  author={Antol, Stanislaw and Agrawal, Aishwarya and Lu, Jiasen and Mitchell, Margaret and Batra, Dhruv and Zitnick, C Lawrence and Parikh, Devi},
  booktitle={Proceedings of the IEEE international conference on computer vision},
  pages={2425--2433},
  year={2015}
}

@inproceedings{goyal2017making,
  title={Making the v in vqa matter: Elevating the role of image understanding in visual question answering},
  author={Goyal, Yash and Khot, Tejas and Summers-Stay, Douglas and Batra, Dhruv and Parikh, Devi},
  booktitle={Proceedings of the IEEE conference on computer vision and pattern recognition},
  pages={6904--6913},
  year={2017}
}

@inproceedings{li2023superclevr,
  title={Super-clevr: A virtual benchmark to diagnose domain robustness in visual reasoning},
  author={Li, Zhuowan and Wang, Xingrui and Stengel-Eskin, Elias and Kortylewski, Adam and Ma, Wufei and Van Durme, Benjamin and Yuille, Alan L},
  booktitle={Proceedings of the IEEE/CVF conference on computer vision and pattern recognition},
  pages={14963--14973},
  year={2023}
}

@inproceedings{yue2025mmmu,
  title={Mmmu-pro: A more robust multi-discipline multimodal understanding benchmark},
  author={Yue, Xiang and Zheng, Tianyu and Ni, Yuansheng and Wang, Yubo and Zhang, Kai and Tong, Shengbang and Sun, Yuxuan and Yu, Botao and Zhang, Ge and Sun, Huan and others},
  booktitle={Proceedings of the 63rd Annual Meeting of the Association for Computational Linguistics (Volume 1: Long Papers)},
  pages={15134--15186},
  year={2025}
}

@article{dao2022flashattention,
  title={Flashattention: Fast and memory-efficient exact attention with io-awareness},
  author={Dao, Tri and Fu, Dan and Ermon, Stefano and Rudra, Atri and R{'e}, Christopher},
  journal={Advances in neural information processing systems},
  volume={35},
  pages={16344--16359},
  year={2022}
}

@inproceedings{xu2024llavacot,
  title={Llava-cot: Let vision language models reason step-by-step},
  author={Xu, Guowei and Jin, Peng and Wu, Ziang and Li, Hao and Song, Yibing and Sun, Lichao and Yuan, Li},
  booktitle={Proceedings of the IEEE/CVF International Conference on Computer Vision},
  pages={2087--2098},
  year={2025}
}

@inproceedings{zhang2025improve,
  title={Improve vision language model chain-of-thought reasoning},
  author={Zhang, Ruohong and Zhang, Bowen and Li, Yanghao and Zhang, Haotian and Sun, Zhiqing and Gan, Zhe and Yang, Yinfei and Pang, Ruoming and Yang, Yiming},
  booktitle={Proceedings of the 63rd Annual Meeting of the Association for Computational Linguistics (Volume 1: Long Papers)},
  pages={1631--1662},
  year={2025}
}

@inproceedings{suris2023vipergpt,
  title={Vipergpt: Visual inference via python execution for reasoning},
  author={Sur{\'\i}s, D{\'\i}dac and Menon, Sachit and Vondrick, Carl},
  booktitle={Proceedings of the IEEE/CVF international conference on computer vision},
  pages={11888--11898},
  year={2023}
}

@inproceedings{mitra2024compositional,
  title={Compositional chain-of-thought prompting for large multimodal models},
  author={Mitra, Chancharik and Huang, Brandon and Darrell, Trevor and Herzig, Roei},
  booktitle={Proceedings of the IEEE/CVF Conference on Computer Vision and Pattern Recognition},
  pages={14420--14431},
  year={2024}
}

@article{chen2023shikra,
  title={Shikra: Unleashing multimodal llm's referential dialogue magic},
  author={Chen, Keqin and Zhang, Zhao and Zeng, Weili and Zhang, Richong and Zhu, Feng and Zhao, Rui},
  journal={arXiv preprint arXiv:2306.15195},
  year={2023}
}

@article{peng2023kosmos,
  title={Kosmos-2: Grounding multimodal large language models to the world},
  author={Peng, Zhiliang and Wang, Wenhui and Dong, Li and Hao, Yaru and Huang, Shaohan and Ma, Shuming and Wei, Furu},
  journal={arXiv preprint arXiv:2306.14824},
  year={2023}
}

@inproceedings{zhao2025swift,
  title={Swift: a scalable lightweight infrastructure for fine-tuning},
  author={Zhao, Yuze and Huang, Jintao and Hu, Jinghan and Wang, Xingjun and Mao, Yunlin and Zhang, Daoze and Jiang, Zeyinzi and Wu, Zhikai and Ai, Baole and Wang, Ang and others},
  booktitle=AAAI,
  volume={39},
  number={28},
  pages={29733--29735},
  year={2025}
}

@article{dao2023flashattention2,
  title={Flashattention-2: Faster attention with better parallelism and work partitioning},
  author={Dao, Tri},
  journal={arXiv preprint arXiv:2307.08691},
  year={2023}
}

@article{loshchilov2017decoupled,
  title={Decoupled weight decay regularization},
  author={Loshchilov, Ilya and Hutter, Frank},
  journal={arXiv preprint arXiv:1711.05101},
  year={2017}
}

@inproceedings{rasley2020deepspeed,
  title={Deepspeed: System optimizations enable training deep learning models with over 100 billion parameters},
  author={Rasley, Jeff and Rajbhandari, Samyam and Ruwase, Olatunji and He, Yuxiong},
  booktitle={Proceedings of the 26th ACM SIGKDD international conference on knowledge discovery \& data mining},
  pages={3505--3506},
  year={2020}
}

@inproceedings{kwon2023efficient,
  title={Efficient Memory Management for Large Language Model Serving with PagedAttention},
  author={Woosuk Kwon and Zhuohan Li and Siyuan Zhuang and Ying Sheng and Lianmin Zheng and Cody Hao Yu and Joseph E. Gonzalez and Hao Zhang and Ion Stoica},
  booktitle={Proceedings of the ACM SIGOPS 29th Symposium on Operating Systems Principles},
  year={2023}
}

@article{sutton1999policy,
  title={Policy gradient methods for reinforcement learning with function approximation},
  author={Sutton, Richard S and McAllester, David and Singh, Satinder and Mansour, Yishay},
  journal={Advances in neural information processing systems},
  volume={12},
  year={1999}
}
\bibliographystyle{ieee_fullname}


\appendix
\clearpage

\setcounter{figure}{0}
\setcounter{table}{0}
\setcounter{equation}{0}
\setcounter{algorithm}{0}
\section*{Appendix}

\section{Implementation Details}
\label{sec:implementation}
This section provides additional implementation details for our experiments.
For clarity, we organize the experiments into five task settings that share the same RLVR training framework but differ in data sources and hyperparameter configurations:
\begin{itemize}
    \item Task 1: Geometry and logic reasoning. Experiments on Geometry3K, MathVista, MathVerse, and LogicVista.
    \item Task 2: Visual grounding. Experiments on RefCOCO and LISA-Grounding.
    \item Task 3: Few-shot classification. Experiments on FGVC Aircraft and Flower102 under 1/2/4-shot settings.
    \item Task 4: Visual puzzle reasoning. Experiments on PuzzleVQA and AlgoPuzzleVQA.
    \item Task 5: Scalability analysis. Experiments on ViRL39K and the multi-benchmark evaluation used for the scaling experiments in the main paper.
\end{itemize}
Across all settings, we adopt a unified RLVR setup. 
As summarized in \tabref{tab:hyperparameters}, we use AdamW~\cite{loshchilov2017decoupled} with full-parameter fine-tuning in bfloat16 precision and gradient checkpointing. 
For each question, we sample 8 responses with temperature $1.0$ and top-$p=1.0$, and train with DeepSpeed ZeRO-2~\cite{rasley2020deepspeed} on 8 NVIDIA A40 GPUs. 
All RLVR methods are implemented within the Swift framework~\cite{zhao2025swift}. 
The gate strength $\alpha$ is constrained to a small shared set across tasks: $\alpha=0.1$ for Tasks 1/4/5, $\alpha=0.05$ for Task 2, and $\alpha=0.02$ for Task 3.
\begin{table}[!htbp]
    \centering
    \caption{Key hyperparameters across the five tasks.}
    \label{tab:hyperparameters}
    \small
    \setlength{\tabcolsep}{20pt}
    \begin{tabularx}{\linewidth}{lll}
        \toprule
        \textbf{Hyperparameter} & \textbf{Task(s)} & \textbf{Value} \\
        \midrule
        \multicolumn{3}{l}{\textit{Shared RL configuration}} \\
        Optimizer & 1--5 & AdamW \\
        Train type & 1--5 & full \\
        Precision & 1--5 & bfloat16 \\
        Gradient checkpointing & 1--5 & Enabled \\
        Responses per question & 1--5 & 8 \\
        Temperature & 1--5 & 1.0 \\
        Top-p & 1--5 & 1.0 \\
        DeepSpeed stage & 1--5 & ZeRO-2 \\
        Iterations & 1--5 & 1 \\
        Freeze Vision Tower & 1--5 & True \\
        \midrule
        \multicolumn{3}{l}{\textit{Optimization hyperparameters}} \\
        Learning rate & 1, 4, 5 & $1 \times 10^{-6}$ \\
        Learning rate & 2, 3 & $2 \times 10^{-6}$ \\
        LR schedule & 1--4 & Cosine \\
        LR schedule & 5 & Constant \\
        Per-device train batch size & 1, 3 & 2 \\
        Per-device train batch size & 2, 4, 5 & 4 \\
        Gradient accumulation steps & 1 & 4 \\
        Gradient accumulation steps & 2, 3, 4 & 2 \\
        Gradient accumulation steps & 5 & 32 \\
        Epochs & 1, 2, 4 & 1 \\
        Epochs & 3 & 4 \\
        Epochs & 5 & 2 \\
        KL coefficient $\beta$ & 1--4 & 0.001 \\
        KL coefficient $\beta$ & 5 & 0.01 \\
        \midrule
        \multicolumn{3}{l}{\textit{Context length and generation}} \\
        Use vLLM~\cite{kwon2023efficient} & 1--5 & True \\
        Attention implementation & 1--5 & FlashAttention2~\cite{dao2023flashattention2} \\
        Max completion length & 1, 5 & 1024 \\
        Max completion length & 2, 3, 4 & 512 \\
        \midrule
        \multicolumn{3}{l}{\textit{DAPO specific hyperparameters}} \\
        Clip Ratio Low & 1--5 & 0.2 \\
        Clip Ratio High & 1--5 & 0.28 \\
        Loss Averaging Mode & 1--5 & Token-level \\
        Max Resample Times & 1--5 & 3 \\
        \midrule
        \multicolumn{3}{l}{\textit{High-entropy RL specific hyperparameters}} \\
        Top Entropy Quantile & 1--5 & 0.2 \\
        \midrule
        \multicolumn{3}{l}{\textit{PEPO specific hyperparameters}} \\
        Gate alpha & 4, 5 & 0.1 \\
        Gate alpha & 1, 2 & 0.05 \\
        Gate alpha & 3 & 0.02 \\
        \bottomrule
    \end{tabularx}
\end{table}

\section{Prompt and Reward Design}
\label{sec:prompt}
The design of prompts and verifiable rewards plays a central role in RLVR, as it governs both how the model articulates its reasoning and how reliable responses can be evaluated.
\tabref{tab:Prompt} summarizes the prompt formats and rewards used across the five task settings.
All tasks employ simple, programmatically verifiable reward components to ensure deterministic and reproducible evaluation.
The format reward verifies whether a response adheres strictly to the prescribed output template (e.g., correct use of \texttt{<think>} / \texttt{<answer>} tags, JSON format for visual grounding, and the \texttt{\textbackslash boxed\{\}} wrapper in Task~5), and assigns a value of 1 only when all format checks are satisfied.
The accuracy reward is binary: the final prediction is extracted from the \texttt{<answer>} span (Tasks~1, 3, 4) or from the content within \texttt{\textbackslash boxed\{\}} (Task~5), and correctness is determined using ground-truth annotations or the official evaluation script.
For visual grounding (Task~2), the IoU reward is given directly as the intersection-over-union value between the predicted box and the best-matching ground-truth box.
The overall rewards for each sample are the weighted sum of these components, as specified in \tabref{tab:Prompt}.
To ensure unambiguous evaluation, all prompts instruct the model to output its intermediate reasoning within \texttt{<think>}~\dots~\texttt{</think>} before producing a concise answer in a designated format (either \texttt{<answer>}~\dots~\texttt{</answer>} or \texttt{\textbackslash boxed\{\}}), which is used exclusively for reward computation.

\begin{table}[!t]
    \centering
    \caption{Prompt and reward design across the five tasks.}
    \label{tab:Prompt}
    \small
    \begin{tabularx}{\linewidth}{l X}
        \toprule
        \textbf{Task(s)} & \textbf{Design} \\
        \midrule
        \multicolumn{2}{l}{\textit{Reward Design}} \\
        1, 3, 4 & Format reward + Accuracy reward \\
        2 & Format reward + IoU reward \\
        5 & 0.1 $\times$ Format reward + 0.9 $\times$ Accuracy reward \\
        \midrule
        \multicolumn{2}{l}{\textit{Prompt Design}} \\
        1 & First output the thinking process in \texttt{<think>} ... \texttt{</think>} tags and then output the final answer in \texttt{<answer>} ... \texttt{</answer>} tags. \\
        2 & First output the thinking process in \texttt{<think>} ... \texttt{</think>} tags and then output the final answer in \texttt{<answer>} ... \texttt{</answer>} tags. Output the final answer in JSON format. \\
        3 & This is an image containing an aircraft/plant. Please identify the model of the aircraft/plant based on the image. Output the thinking process in \texttt{<think>} ... \texttt{</think>} and the final answer in \texttt{<answer>} ... \texttt{</answer>} tags. The output format should be: \texttt{<think> ... </think> <answer>species name</answer>}. Please strictly follow the format. \\
        4 & First output the thinking process in \texttt{<think>} ... \texttt{</think>} tags and then output the final answer in \texttt{<answer>} ... \texttt{</answer>} tags. Answer with the option letter directly. \\
        5 & First think through the reasoning process as an internal monologue, enclosed within \texttt{<think>} ... \texttt{</think>} tags. Then provide the final answer enclosed within \texttt{\textbackslash boxed\{\}}. \\
        \bottomrule
    \end{tabularx}
\end{table}

\section{Evaluation}
\label{sec:evaluation}
We adopt the standard dataset splits and employ deterministic, programmatic evaluation.

\myPara{Geometry and logic reasoning.}
For Geometry3K (validation and test), MathVista-mini, MathVerse-mini, and LogicVista, we report accuracy averaged over multiple samples.
For each image--question pair, the model generates 8 responses with temperature set to 1.0, and computes the avg@8 accuracy.
To avoid any LLM-as-a-judge evaluation and ensure that correctness is decided purely by programmatic rules, we omit instances with free-form answers on all geometry reasoning out-of-domain datasets, except MathVista-mini, which is evaluated strictly using the official script released by the authors, including their normalization and correctness checker.

\myPara{Few-shot classification.}
For FGVC Aircraft and Flower102, we follow the 1-, 2-, and 4-shot settings described in the experiment setup section of the main paper (Sec.~4.1).
The support images and labels are included in the prompt, and the model predicts the class name of the query image via greedy decoding.
A prediction is considered correct if the decoded class name exactly matches the ground-truth category label.

\myPara{Visual puzzle reasoning.}
For PuzzleVQA and AlgoPuzzleVQA, we report top-1 accuracy.
Each question is associated with a predefined set of options, and the model outputs a single option label enclosed in \texttt{<answer>} ... \texttt{</answer>} tags.
We use greedy decoding, and any malformed or ambiguous output is treated as incorrect.

\myPara{Visual grounding.}
For RefCOCO and LISA-Grounding, we follow the standard phrase grounding protocol and report IoU@50.
The model predicts a single bounding box using greedy decoding (temperature 0).
A prediction is deemed correct if the predicted bounding box attains an IoU of at least 0.5 with any ground-truth annotation.

\myPara{Scaling benchmarks.}
For the ViRL39K scalability experiments, models are trained on ViRL39K and evaluated on Geometry3K\textsubscript{test}, MathVista, We-Math, MathVerse, LogicVista, SuperClevr Counting, and MMMU-Pro.
We employ the public evaluation scripts from PAPO to ensure fair comparison.
Following PAPO, instances requiring an LLM-as-a-judge are excluded for all datasets except Geometry3K, ensuring that all reported metrics rely solely on deterministic checkers.
All benchmarks adopt the same avg@8 protocol described above.

\begin{algorithm}[t]
\caption{Perception--Exploration Policy Optimization (PEPO)}
\label{alg:pepo}
\begin{algorithmic}[1]
\REQUIRE Policy $\pi_\theta$, old policy $\pi_{\text{old}}$, group size $G$, maximum steps $K_{\max}$, gate strength $\alpha$, learning rate $\eta$
\FOR{$k = 1,\dots,K_{\max}$}
    \STATE Sample prompts and grouped responses $\{\tau^{(i)}\}_{i=1}^{G}$ from $\pi_{\text{old}}$
    \STATE Compute rewards $R^{(i)}$ and GRPO advantages
    \STATE $A^{(i)} \gets \dfrac{R^{(i)} - \mathrm{mean}_j R^{(j)}}{\mathrm{std}_j R^{(j)} + \varepsilon}$
    \STATE \textbf{Perception modeling (visual similarity):}
    \STATE For each token $t$ in response $i$, compute
    \STATE $\displaystyle
    \mathrm{VS}_t^{(i)} \gets
    \frac{1}{L} \sum_{l=1}^{L} \frac{1}{N} \sum_{n=1}^{N}
    \frac{\langle h_{l,t}^{(i)}, v_{l,n}^{(i)} \rangle}
         {\|h_{l,t}^{(i)}\| \,\|v_{l,n}^{(i)}\|}$
    \STATE \textbf{Exploration modeling (entropy):}
    \STATE For each token $t$ in response $i$, compute
    \STATE $\displaystyle
    H_t^{(i)} \gets
    - \sum_{x \in \mathcal V}
    p_\theta(x \mid s_t^{(i)})
    \log p_\theta(x \mid s_t^{(i)})$
    \STATE \textbf{Perception--exploration fusion:}
    \STATE Min--max normalize $\mathrm{VS}_t^{(i)}$ and $H_t^{(i)}$ over $t$ to obtain $\hat{\mathrm{VS}}_t^{(i)}, \hat H_t^{(i)} \in [0,1]$
    \STATE $\hat g_t^{(i)} \gets \hat{\mathrm{VS}}_t^{(i)} + \hat H_t^{(i)} - \mathrm{mean}_t(\hat{\mathrm{VS}}^{(i)} + \hat H^{(i)})$
    \STATE $\displaystyle
    w_t^{(i)} \gets
    T \cdot \mathrm{Softmax}_t\big((1 + \alpha \tanh(\hat g_t^{(i)})) \,\mathrm{VS}_t^{(i)}\big)$
    \STATE \textbf{Token-level advantage:}
    \STATE $\lambda_k \gets \min(1,\, k / K_{\max})$
    \STATE $A_t^{(i)} \gets \big[(1-\lambda_k) + \lambda_k w_t^{(i)}\big]\, A^{(i)}$
    \STATE \textbf{Policy update:}
    \STATE Use $A_t^{(i)}$ in place of $A^{(i)}$ in a standard GRPO / PPO-style objective $J(\theta)$
    \STATE $\theta \gets \theta + \eta \nabla_\theta J(\theta)$
    \STATE $\pi_{\text{old}} \gets \pi_\theta$
\ENDFOR
\end{algorithmic}
\end{algorithm}

\section{Policy Gradient View of PEPO}
\label{sec:appendix_policy_gradient}
In \algref{alg:pepo}, \methodname\ introduces token-wise weights $\{w_t^{(i)}\}_{t=1}^{T}$ to refine the sequence-level advantage of GRPO.
We show that the unit-mean constraint on $w_t^{(i)}$ preserves the sequence-level policy-gradient scale and only redistributes credit among tokens.

\myPara{Unit-mean property.}
For a response of length $T$ we set
\begin{equation}
    w_t^{(i)} = T \cdot \mathrm{Softmax}\!\big(z_t^{(i)}\big),
\end{equation}
so that
\begin{equation}
    \frac{1}{T}\sum_{t=1}^{T} w_t^{(i)}
    = \sum_{t=1}^{T} \mathrm{Softmax}\!\big(z_t^{(i)}\big)
    = 1
    \Longrightarrow
    \sum_{t=1}^{T} w_t^{(i)} = T.
    \label{eq:sum_w_equals_T}
\end{equation}

\myPara{Effect on the policy gradient.}
For response $i$, the (unclipped) GRPO policy gradient~\cite{sutton1999policy} can be written as
\begin{equation}
    \nabla_\theta J_{\text{G}}(\theta)
    = \mathbb{E}\left[
    \sum_{t=1}^{T}
    A^{(i)}\,
    \nabla_\theta
    \log \pi_\theta\big(x \mid s_t^{(i)}\big)
    \right],
    \label{eq:pg_grpo_short}
\end{equation}
where $A^{(i)}$ is the sequence-level advantage.
In \methodname, we use token-wise advantages
\begin{equation}
    A_t^{(i)} = \big[(1-\lambda) + \lambda w_t^{(i)}\big] A^{(i)}.
\end{equation}
The total advantage per sequence becomes
\begin{align}
    \sum_{t=1}^{T} A_t^{(i)}
    &= A^{(i)} \sum_{t=1}^{T}
       \big[(1-\lambda) + \lambda w_t^{(i)}\big] \nonumber \\
    &= A^{(i)}\!\big[(1-\lambda)T + \lambda \sum_{t=1}^{T} w_t^{(i)}\big]
       \nonumber \\
    &= A^{(i)} T,
    \label{eq:sequence_mass_preserved}
\end{align}
Therefore, \methodname{} preserves the total sequence-level advantage mass $T A^{(i)}$ and does not introduce a global scaling change in the policy gradient.
The modification affects only the distribution of credit across tokens within the summation in \eqnref{eq:pg_grpo_short}.
As a result, the overall gradient magnitude remains consistent with GRPO, while credit is preferentially allocated to visually grounded or high-entropy tokens.

\section{Additional Ablation Results}
\label{sec:appendix_ablation}

The main paper examines the roles of perception and exploration in \methodname{} and the key elements of its weighting strategy. Here we provide additional ablations to further characterize the method under different settings.

\begin{table*}[!t]
  \centering
  \small
  \setlength{\tabcolsep}{9pt}
  \caption{Ablation on gate strength $\alpha$ of \methodname\ on geometry and logic reasoning benchmarks using the Qwen2.5-VL-3B-Instruct model.}
  \label{tab:alpha_ablation_geo}
  \resizebox{0.99\textwidth}{!}{
  \begin{tabular}{lcccccc}
    \toprule
    $\alpha$
    & \textbf{Geometry3K}\textsubscript{val}
    & \textbf{Geometry3K}\textsubscript{test}
    & \textbf{MathVista}\textsubscript{mini}
    & \textbf{MathVerse}\textsubscript{mini}
    & \textbf{LogicVista}
    & \textbf{Avg} \\
    \midrule
    0.00 (Perception-only)
    & 21.07 & 26.73 & 52.81 & 45.31 & 34.93 & 36.17 \\
    0.02
    & 20.60 & 25.87 & 53.88 & 44.99 & 34.73 & 36.01
    \\
    \rowcolor[HTML]{EFEFEF} 0.05 (Default)
    & 21.91 & 27.27 & 54.45 & 45.42 & 34.45 & 36.70 \\
    0.10
    & 22.80 & 26.81 & 53.58 & 44.12 & 34.26 & 36.31 \\
    0.15
    & 21.45 & 26.52 & 53.54 & 43.99 & 34.62 & 36.02  \\
    \bottomrule
  \end{tabular}
  }
\end{table*}

\myPara{Effect of gate strength $\alpha$.}
To investigate how entropy interacts with visual similarity, we vary the gate strength $\alpha$ in the perception--exploration fusion module.
As shown in \tabref{tab:alpha_ablation_geo}, \methodname{} remains robust across a range of $\alpha$ values.
The perception-only variant ($\alpha=0$) already outperforms GRPO, indicating that visual similarity alone provides a strong training signal.
Adding a moderate entropy gate further improves performance, with the best average results obtained at $\alpha=0.05$ or $\alpha=0.10$ depending on the benchmark.
When $\alpha$ is chosen either too small or too large, the overall performance drops slightly.
This trend suggests that entropy is most beneficial when it serves as a complementary modulation signal rather than being completely removed or overly amplified.
Overall, the results support both the robustness of \methodname{} to $\alpha$ and the effectiveness of using a modest gate strength.

\myPara{Effect of perception prior measure.}
We analyze the impact of different similarity measures used to compute the perception prior in \methodname. 
Our default choice is the cosine similarity between hidden states and visual embeddings, motivated by its scale invariance and its widespread effectiveness in vision--language alignment (e.g., CLIP-style representations).
We compare cosine similarity with two distance-based alternatives, namely L1 and L2 distances.
As shown in \tabref{tab:sim_metric_ablation}, cosine similarity consistently achieves better performance across geometry reasoning and few-shot classification benchmarks.
In particular, replacing cosine similarity with L1 or L2 distance leads to noticeable performance degradation, indicating that cosine similarity provides a more stable and semantically aligned signal for perception modeling.

\begin{table}[t]
\centering
\caption{Ablation on perception prior measures used to compute visual similarity.}
\label{tab:sim_metric_ablation}
\small
\setlength{\tabcolsep}{15pt}
\begin{tabular}{lccc}
\toprule
\textbf{Measure} & \textbf{Geometry3K$_{val}$} & \textbf{Geometry3K$_{test}$} & 
\textbf{FGVC Aircraft (4-shot)} \\
\midrule
L1 distance & 19.00 & 24.50 & 60.85 \\
L2 distance & 20.90 & 26.08 & 67.33 \\
\rowcolor[HTML]{EFEFEF} Cosine similarity (Default) & 22.80 & 26.81 & 75.79 \\
\bottomrule
\end{tabular}
\end{table}

\section{Additional Details for the Hidden-state Shift Analysis}
\label{sec:Representation-Shift}

\myPara{Hidden-state shift across tokens.}
The bar plot in the main paper summarizes the hidden-state shift of response tokens when grouped into bins defined by their visual similarity or token-level entropy.
Given hidden states $h^{\text{with}}_{l,t}$ and $h^{\text{without}}_{l,t}$ for token $t$ at layer $l$ with and without the image input, respectively, the representational shift associated with token $t$ is defined as
\begin{equation}
D_t = \frac{1}{L}\sum_{l=1}^{L}\big\|h^{\text{with}}_{l,t} - h^{\text{without}}_{l,t}\big\|_2.
\end{equation}
Tokens are then assigned to percentile bins according to either their visual similarity $\mathrm{VS}_t$ or their entropy, both computed under the image-present condition, and we report the average value of $D_t$ within each bin.

\myPara{Word cloud construction.}
The word clouds in this analysis illustrate lexical patterns associated with high-entropy and high-visual-similarity tokens.
To construct these word clouds, we aggregate per-token statistics over Geometry3K as follows:
\begin{itemize}
  \item Tokens with fewer than 50 occurrences are excluded to avoid instability due to rare or noisy items.
  \item For entropy-based clouds, tokens are ranked by their mean token-level entropy; for perception-based clouds, they are ranked by their mean visual similarity $\mathrm{VS}_t$.
  \item From each ranking, we select a fixed set of the top 100 tokens and weight them by their aggregated frequencies when rendering the clouds.
  \item Special tokens and non-semantic artifacts (e.g., control tokens and markup) are removed.
\end{itemize}

\section{Limitations}
\label{sec:limitations}
Although we demonstrate the effectiveness of \methodname{} on recent 2B or 3B LVLMs under several RLVR training pipelines, we do not extend our experiments to larger model backbones (e.g., 7B or above) or to longer-context configurations due to computational and memory limitations.
Furthermore, our evaluation is confined to a curated set of multimodal reasoning and grounding benchmarks.
Applying \methodname{} to stronger base models and to a broader range of tasks, such as video understanding and tool-augmented reasoning, is an important direction for future research.

\end{document}